\pdfoutput=1

\documentclass[11pt]{article}

\usepackage[]{ACL2023}

\usepackage{times}
\usepackage{latexsym}

\usepackage[T1]{fontenc}

\usepackage[utf8]{inputenc}

\usepackage{microtype}

\usepackage{inconsolata}
\usepackage{graphicx,adjustbox,booktabs,arydshln}
\usepackage{adjustbox}
\usepackage{amsmath}

\newenvironment{sizeddisplay}[1]
 {\par\nopagebreak#1\noindent\ignorespaces}
 {\nopagebreak\ignorespacesafterend}

\title{Data-Efficient Finetuning Using Cross-Task Nearest Neighbors}

\author{Hamish Ivison$^{\alpha}$ \quad Noah A. Smith$^{\alpha\beta}$ \quad Hannaneh Hajishirzi$^{\alpha\beta}$ \quad Pradeep Dasigi$^{\alpha}$ \quad
\\
$^{\alpha}$Allen Institute for AI\\
$^{\beta}$Paul G.~Allen School of Computer Science \& Engineering, University of Washington\\
\texttt{\{hamishi,noah,hannah,pradeepd\}@allenai.org} \\
}

\begin{document}
\maketitle
\begin{abstract}
Obtaining labeled data to train a model for a task of interest is often expensive. Prior work shows training models on multitask data augmented with task descriptions (prompts) effectively transfers knowledge to new tasks. Towards efficiently building task-specific models, we assume access to a small number (32--1000) of unlabeled target-task examples and use those to retrieve the most similar labeled examples from a large pool of multitask data augmented with prompts. Compared to the current practice of finetuning models on uniformly sampled prompted multitask data (e.g., FLAN, T0), our approach of finetuning on cross-task nearest neighbors is significantly more data-efficient. Using only 2\% of the data from the P3 pool without any labeled target-task data, our models outperform strong baselines trained on all available data by 3--30\% on 12 out of 14 datasets representing held-out tasks including legal and scientific document QA. Similarly, models trained on cross-task nearest neighbors from SuperNaturalInstructions, representing about 5\% of the pool, obtain comparable performance to state-of-the-art models on 12 held-out tasks from that pool. Moreover, the models produced by our approach also provide a better initialization than single multitask finetuned models for few-shot finetuning on target-task data, as shown by a 2--23\% relative improvement over few-shot finetuned T0-3B models on 8 datasets. We publicly release our code.\footnote{{\scriptsize\texttt{\href{https://github.com/allenai/data-efficient-finetuning}{https://github.com/allenai/data-efficient-finetuning}}}}

\end{abstract}

\section{Introduction}

\begin{figure}[]
    \centering
    \adjustbox{max width=.5\textwidth}{
        \includegraphics{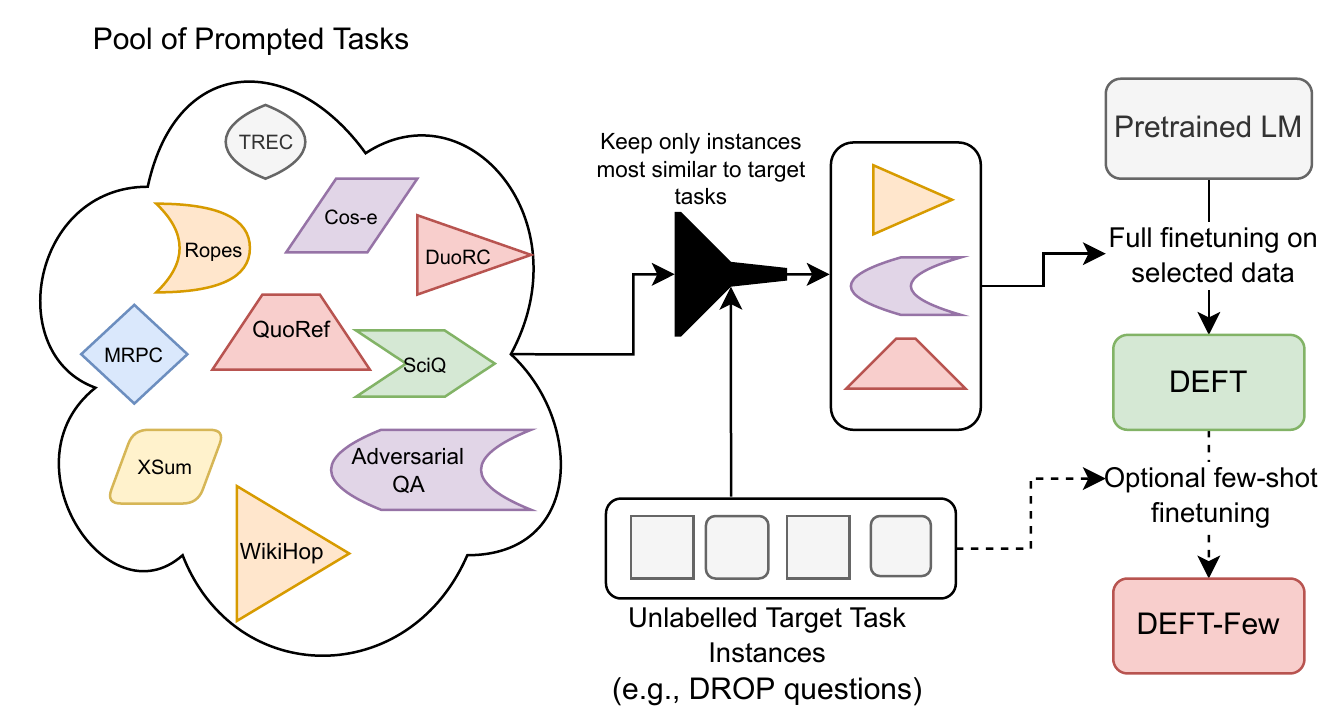}
    }
    \caption{Overview of the DEFT method. Given some unlabeled target-task instances, we find the most similar instances in a large pool of multitask data. We train a model on these instances. If we have access to labeled data, we optionally few-shot finetune the DEFT model.}
    \label{fig:retrieve}
\end{figure}
Finetuning large models with data from a diverse set of tasks, augmented to include brief descriptions of the tasks (i.e., prompts) has been shown to help models generalize to unseen tasks \citep{wei2021finetuned,sanh2021multitask}. This cross-task generalization capability is particularly helpful in cases where it is expensive to collect labeled target task training sets. Prior work trained single models with as much prompted data as possible --- for example, \citet{sanh2021multitask} train a model on roughly 11 million instances (counting different prompt variations). The training datasets were selected without using any information about the target tasks, with the goal of allowing models to generalize to new tasks from instructions alone, making the evaluation ``zero-shot''. However, it is unclear if all the training data is required for good performance on any given single target task. Furthermore, given that neural network models have previously been shown to suffer from negative interference (wherein training on more datasets results in worse performance on certain downstream tasks) in multitask setups \citep{aribandi2022ext} and benefit from pretraining on domain-relevant data \citep{gururangan-etal-2020-dont, Phang2018SentenceEO}, it is possible that training only on relevant prompted data could further improve task generalization while being data-efficient.

Based on this hypothesis, we seek to make use of unlabelled data to find relevant subsets of training data in the massive pool of multitask data, allowing similar-to-better performance than training on the entire pool for a given target task. Manually finding relevant training data in a massive pool of data is infeasible since it is not obvious which of the source tasks are relevant for a given target task, and which instances are most relevant for target task generalization within a source task dataset (see Section~\ref{sec:heatmap}). Hence we rely on a simple method to \textit{automatically} select these subsets. Additionally, as only some samples within a given dataset may be relevant to a target task, we select per-instance rather than per-dataset, unlike prior work, which tries to identify useful datasets for transfer learning \citep{aribandi2022ext, Phang2018SentenceEO} and train on all data within the chosen datasets. We use a setup similar to work examining retrieval-augmented cross-task generalization \citep{Lin2022UnsupervisedCG}: we assume access to a small number of \textit{unlabeled} target task instances and use these to retrieve \textit{cross-task nearest neighbors}---labeled instances from the massive pool of data most similar to our unlabeled target task instances. The similarity is computed as the distance between the representations produced by the encoder of a pretrained seq2seq model. Unlike prior work, we then finetune target task specific models on these neighbors alone, without using any target task specific labeled data or any extra data from the pool of multitask data.
We hope that the similarity between the cross-task neighbors and our target task data will enable better generalization to our target task, with dissimilar examples that may cause interference removed from the training mixture. We ultimately aim to produce models that perform at least as well as models trained on the entire multitask pool \textit{despite being trained on a fraction of data}, greatly reducing the cost of training through the use of a few cheap-to-collect unlabelled examples.

We run experiments with T5~\citep{raffel2020exploring} models, and use Public Pool of Prompts (P3;~\citealp{sanh2021multitask}) as the main pool of prompted multitask data from which to retrieve cross-task nearest neighbors. We evaluate on the 11 datasets originally used to evaluate T0 (a collection of natural language understanding and commonsense tasks), as well as 3 additional datasets with varied domains (e.g., legal, NLP domains). We also experiment with the train set of SuperNaturalInstructions \citep[SNI;][]{niv2} as a pool of multitask data, and evaluate on 12 tasks from SNI's held-out set of test tasks. Our findings are as follows:
\begin{itemize}
    \item For 12 out of 14 target datasets, we find that their cross-task nearest neighbors, at most 2\% of instances retrieved from P3, are much more relevant as training data than the rest of the P3 pool---training T5 models, sometimes even variants smaller than T0-3B, on these subsets yields models with performance 3--30\% better than T0-3B evaluated zero-shot. Similarly, models trained on cross-task neighbors in SuperNaturalInstructions (at most 5\% of the pool), perform similarly to state-of-the-art models trained on all available data. 
    \item For some target tasks on which T0-3B performs close to random chance, T5 models of the same size trained using cross-task nearest neighbors perform significantly above chance, supporting our hypothesis that massive multitask prompted training could lead to negative interference between tasks.
    \item When target task labeled data is available for few-shot finetuning, we find that T5 models trained with cross-task nearest neighbors provide better initialization for parameter-efficient finetuning methods than T0-3B, performing 2--23\% better than T0-3B with few-shot finetuning across 10 out of 11 datasets.
    \item An analysis of what relevant data gets retrieved shows that most of the tasks in the massive pool of multitask data are not retrieved for any target tasks, confirming our hypothesis that only a small subset of data within the pool is relevant to any given target task.
    \item We compare model performance from DEFT with that from full-finetuning across a variety of labelling budgets and find that DEFT is more effective for smaller labelling budgets.
\end{itemize}

These findings suggest that instead of training single models on all available data, multi-task data can be used much more efficiently towards improving model performance on specific target tasks by selecting training data relevant to those tasks, even with a simple method for identifying such data.

\section{Related Work}
\paragraph{Multi-task transfer  models} Training on large multi-task mixtures is a common trend within NLP, with most existing approaches first training a pretrained language model on a large collection of tasks, and then evaluating these models in either zero- or few-shot settings on a collection of held-out datasets \citep{wei2021finetuned, sanh2021multitask, khashabi-etal-2020-unifiedqa, natint, aribandi2022ext}. Most approaches do not customise their task selection to downstream tasks and assume no knowledge of the target tasks ahead of time, instead focusing on building a single model most applicable to any arbitrary evaluation task. In contrast, we show that if we assume access to unlabeled target task instances, we can make much better use of the multitask data, selecting only instances useful to a given task. Relatedly, \citet{vu-etal-2020-exploring} propose a method for using gradients from labelled task data to construct task embeddings for predicting task transferability. Our method instead uses unlabeled data, which is much cheaper and easier to collect, and does not use gradients, making it easier to scale to large models such as T5-XL.

\paragraph{Retrieval-based methods for NLP} Adding retrieval components to language models has been shown~\citep{khandelwal2019generalization,guu2020retrieval,lewis2020retrieval} to augment their generalization capabilities by externalizing memorization. In contrast to prior work in this direction that mostly focused on language modeling as the end task, we evaluate on a variety of language understanding tasks. The work from~\citet{shi2022nearest} used retrieval-based methods for classification tasks by heuristically mapping the label space of the end-tasks to that of the predicted next words of the nearest neighbors from a language model. We instead finetune the models on the nearest neighbors.
\citet{Lin2022UnsupervisedCG} also use unlabeled examples to retrieve relevant data for improving performance but focus on \textit{further finetuning multi-task models}. They use representations from the encoder of a multi-task finetuned model (e.g. T0) to retrieve subsets of its training data closest to the instances of a target dataset and further finetune the model to specialize it for the target task. While their results suggest that  using a multi-task model is crucial for good retrieval performance, we show gains using a model before multitask finetuning. Our setup allows for data-efficiency via \textit{pruning the amount of multi-task data used during training}, letting a practitioner who only cares about specific downstream tasks train strong task-specific models using much less data and compute than if they trained on the entire pool of multi-task data.

\paragraph{Parameter-efficient fine-tuning} In contrast to work that focused on finetuning fewer parameters in large models to adapt them to new tasks~\citep{houlsby2019parameter,hu2021lora,liu2022few}, our proposal is a  \textit{data-}efficient training method for obtaining task-specific models without using target task labels. Our method is complementary to parameter-efficient methods, and they can be used in conjunction, as shown in section~\ref{sec:few_shot_exp}.

\paragraph{Instance attribution} Our approach works by identifying the most relevant training examples for a given data point, which is called \textit{instance attribution}. Prior work~\citep{koh2017understanding,yeh2018representer,pruthi2020estimating,han2022orca} used instance attribution methods to interpret predictions of neural network models. These methods generally relied on the gradients of the model to identify the effect specific data points, either in the pretraining or the finetuning stage, have on the model's predictions. Our method for identifying cross-task neighbors is simpler because we do not use gradients and we do not even rely on the labels of the data. Results from~\citet{pezeshkpour2021empirical} show that instance attribution based on similarity between the model's representations is comparable to gradient-based approaches in terms of finding the most important training data points. 

\paragraph{Making use of auxiliary data} Training on intermediate data has been shown to improve performance on target NLP tasks \citep{phang2018stilts}. Recent work has shown that intermediate datasets can be selected by embedding-based methods \citep{vu-etal-2020-exploring, poth-etal-2021-pre, kung-etal-2021-efficient}. Most prior work relies on expensive embedding computation methods, either training a model to generate task embeddings, or using methods that are difficult to scale to large models.\footnote{E.g., the Fisher information matrix used by \citet{vu-etal-2020-exploring}.} In contrast, we use an extremely cheap embedding method (mean-pooling over an encoder), and additionally consider sample-wise selection over a massive pool of tasks, as opposed to selecting entire tasks.

\section{Data Efficient Finetuning across Multiple Tasks}
Given a large collection of labeled prompted data (i.e., data converted into a text-to-text form, with task instructions included in the input, e.g., P3), our core hypothesis is that some tasks in this massive pool of data are more similar to a given target task than others. Given a target task, we assume we have access to a small amount of \textit{unlabeled} target task data, which is often much easier and cheaper to collect than labeled data (see Section \ref{sec:case_study}). Our aim is to find a relevant subset of data from our pool given a single target task, ideally allowing us to train a model using this subset that outperforms a similar model trained on the entire pool of data.

Manually identifying the relevant subsets of these datasets is not feasible since task boundaries are usually not clearly defined in NLP, and it is hard to interpret what skills get transferred when a model is trained on one dataset and evaluated on other. Hence, we use the similarity between the pretrained model's representations to compute relevance. We encode all instances in the large pool of multitask data with a pretrained language model and build a search index over the resulting representations. Given small amounts of unlabeled target task data, we retrieve relevant multitask subsets from the index, which we call \textbf{cross-task nearest neighbors} of the target tasks. We then build task-specific models by finetuning the pretrained models on the cross-task neighbors. We refer to this approach as \textbf{Data-Efficient FineTuning} (\textbf{DEFT}).

We evaluate our approach both in cases where no labeled data is available, and when a few (20--70) annotated labels are available. In the former case, we simply use the unlabeled data for retrieval and evaluate the resulting DEFT model ``zero-shot'' on the target task. In the latter case, we first train a DEFT model and then perform parameter-efficient few-shot tuning using IA3 \citep{liu2022few} to make use of the labeled data.

\paragraph{Retrieving cross-task nearest neighbors} To retrieve the most similar instances to a given set of target task instances, we first build an index over the massive pool of multi-task data for efficient retrieval, encoding samples using a pretrained encoder. Then, given a set of query instances $Q$, we retrieve our subset of similar data by computing a union of the $k$-nearest neighbors to all $q \in Q$. Note that there may be an overlap between the sets of nearest neighbors retrieved for different queries, and hence $|R| \leq |Q| \cdot k$, where $R$ is the retrieved subset. Empirically, we find $|R|$ tends to be 5--50$\times$ smaller than $|Q| \cdot k$ due to this overlap.

\paragraph{Data-Efficient FineTuning (DEFT)}\label{sec:deft_train} Given a retrieved set of data $R$, we can then finetune a pretrained language model on the mixture of data using a cross-entropy loss, as all data are in a unified text-to-text prompted format. This training is similar to the multitask prompted training of T0~\citep{sanh2021multitask}. We refer to models trained on $R$ as DEFT models. In settings where we have no labeled data available, we directly evaluate these models on our target tasks.

\paragraph{Parameter-efficient few-shot finetuning} For the case where a few annotated labels are available, we make use of parameter-efficient few-shot finetuning. For this, we take our multi-task trained DEFT checkpoints and finetune them using IA3 \citep{liu2022few} on task-specific few-shot data. Concretely, given a trained transformer model, we introduce three vectors $l_\text{k}$, $l_\text{v}$, and $l_\text{ff}$ into the attention and feed-forward mechanisms of each layer:
\begin{sizeddisplay}{\small}
\begin{align} 
    \text{Attn}(Q, K, V) &= \text{softmax}\left(\frac{Q(l_\text{k}\odot K^T)}{\sqrt{d_k}}\right)(l_\text{v} \odot V) \\
    \text{FFN}(x) &= (l_\text{ff} \odot f(W_1x))W_2
\end{align}
\end{sizeddisplay}
We initialize these vectors with all ones and only update them during the few-shot finetuning. This provides an efficient method of further training our DEFT models on task-specific data and has been shown to outperform full finetuning in the few-shot setting \citep{liu2022few}.

\section{Experiments}

\subsection{Setup \& Hyperparameters}

\paragraph{Indexing P3} We construct an index of P3 data using FAISS~\citep{johnson2019billion}, a library for efficient similarity search over dense vectors. We use a Hierarchical Navigable Small World index~\citep{hnsw} to approximate the $k$-nearest neighbor search. To keep the size of the index manageable, we use Product Quantization~\citep{jegou2010product} and reduce the dimensionality of the encoded representations using an optimized product quantization transform~\citep{ge2013optimized}. We encode our instances using the T5 v1.1 model with extra language model pretraining introduced by \citet{lester-etal-2021-power}. For all experiments, we match the size of the encoder used to index data and the size of downstream models trained on this data (e.g., if we train a T5-XL sized model, we use T5-XL to index and retrieve the data). We use the subset of P3 used to train T0 as our pool of multitask data unless otherwise stated.

\paragraph{DEFT} Following T0, we start with the T5 v1.1 model with extra language model pretraining. Unless otherwise stated, we use the `XL' variant with 3 billion parameters across our experiments. When training on cross-task nearest neighbors, we train for 5 epochs with a batch size of 8 using the Adam optimizer~\citep{adam} and a learning rate of 0.00005. We use a linear warmup schedule for the first 10\% of the total training steps and linear decay for the rest of training.

\paragraph{Few-shot training} We follow the settings suggested by \citet{liu2022few}: training for 1000 steps with a batch size of 8. We use the Adafactor optimizer with a maximum learning rate of 0.003 and a linear decay schedule with 60 warmup steps. We only update the IA3 vectors during training.

\paragraph{Evaluation datasets} We evaluate on the set of 11 datasets used to evaluate T0 (RTE, ANLI R1/2/3, CB, HellaSwag, Story Cloze, WinoGrande, WSC, COPA, WiC), which include natural language inference and commonsense reasoning datasets. In addition to the T0 evaluation datasets, we also evaluate on three additional datasets from diverse domains: CaseHold~\citep{chalkidis-etal-2021-lexglue, casehold}, a legal QA dataset, DROP~\citep{dua-etal-2019-drop}, a QA dataset that requires discrete operations, and a subtask of Qasper~\citep{dasigi-etal-2021-dataset}, a QA dataset over entire NLP papers. Qasper has two subtasks---selecting paragraphs in the paper that provide evidence for answering the questions, and generating the answers. We focus on the former because it was shown to be the more difficult of the two, and convert it into a binary classification task where the inputs are combinations of questions and single paragraphs. We refer to this subtask as \textit{QasperEvidence} henceforth and evaluate model performance in terms of document-level F1 as described by~\citet{dasigi-etal-2021-dataset}. For evaluation and few-shot training, we convert all datasets to a prompted text-to-text format\footnote{For example, ANLI instances were converted to  `\{premise\} Question: \{hypothesis\} True, False, or Neither?', with the answers as `true', `false', or `neither'.} either using the prompt templates from P3 for the T0 evaluation datasets or an original prompt for the other datasets. For CaseHold, DROP, and QasperEvidence we randomly split out 1000 examples from the existing validation sets to use for retrieval, and use the remaining data for evaluation. For all other datasets, we retrieve using up to 1000 randomly chosen examples from the training splits (if a dataset has less than 1000 training examples, we use all available training data for retrieval). We provide further details in Appendix \ref{app:dataset_details}.

\paragraph{Model evaluation} Following \citet{sanh2021multitask} and \citet{brown_gpt_3}, we calculate accuracy on all datasets except DROP using \textit{rank classification}, where we pick the answer with lowest loss across possible answer choices given the instance input as the model prediction. As DROP is a QA dataset that requires selecting spans or generating numbers, and does not have answer choices, we generate the prediction using greedy decoding.

\paragraph{Baselines} For zero-shot evaluation, we primarily compare against 4 baselines: 1) \textit{T0-3B}, trained on about 10\% of the P3 data,\footnote{\citet{sanh2021multitask} report that they train T5-XL on at most 500K instances per prompted dataset in P3, which amounts to about 10\% of the pool.} 2) \textit{Random}, a model trained on a random selection of P3 data the same size as the subsets selected by DEFT, 3) \textit{T5-XL} not finetuned any further, and 4) \textit{BM25}, using BM25\footnote{We use Pyserini \citep{Lin_etal_SIGIR2021_Pyserini} with default settings for the BM25 index. We retrieve the same amount of data as the subsets retrieved by DEFT.} \citep{bm25} for retrieval instead of dense representations. For few-shot settings, we compare T0-3B with additional few-shot training with DEFT checkpoints trained on subsets chosen using (a) 1000 unlabeled instances and (b) the instances used in the few-shot training without labels. This means (b) uses no additional data compared to T0-3B with few-shot finetuning.

\subsection{Data-Efficient Fine-Tuning vs. Massive Multitask Training}
\label{sec:exp_cross}

\begin{table*}[]
\centering
\adjustbox{max width=\textwidth}{
\begin{tabular}{@{}lcccccc|ccc|c@{}}
\toprule
\textbf{Task}       & \textbf{DEFT-XL} & \textbf{T0-3B} & \textbf{Rand-XL} & \textbf{Rand-Bal} & \textbf{T5-XL} & \textbf{BM25-XL} & \textbf{DEFT-base} & \textbf{Rand-base} & \textbf{T5-base} & \textbf{Maj. Class} \\ \midrule
CaseHold & 37.2 & 30.9 & 19.0 & \textbf{38.7} & 11.4 & 27.9 & \textbf{18.9} & 17.5 & 11.4 & 6.6	\\
DROP           & \textbf{31.0} & 27.4 & 24.3 & 27.6 & 11.3 & 22.6 & \textbf{21.3} & 18.0 & 4.0 & - \\
QasperEv. & \textbf{28.5} & 19.9 & 17.9 & 23.2 & 8.2 & 20.3 & \textbf{15.9} & 11.0 & 8.2 & 19.9 \\ \hdashline
RTE                 & 74.0 & 70.4 & \textbf{78.3} & 78.0 & 53.1 & 74.3 & \textbf{61.7} & 61.0 & 52.0 & 53.4 \\
ANLI R1             & 39.8 & 35.0 & 35.3 & \textbf{40.0} & 32.9 & 37.5 & 29.6 & \textbf{33.3} & 32.9 & 33.4 \\
ANLI R2             & \textbf{37.5} & 32.6 & 35.3 & 36.9 & 33.5 & 36.9 & 32.5 & 22.3 & \textbf{33.5} & 33.4 \\
ANLI R3             & 41.4 & 35.3 & 38.0 & \textbf{41.7} & 33.8 & 41.1 & 31.6 & \textbf{33.1} & 32.7 & 33.5 \\
CB                  & \textbf{60.7} & 58.9 & \textbf{60.7} & 55.4 & 44.6 & 50.0 & \textbf{50.0} & 48.2 & 44.6 & 50.0 \\
HellaSwag           & \textbf{33.1} & 28.2 & 27.4 & 29.3 & 23.0 & 28.7 & \textbf{25.9} & 25.0 & 23.0 & 25.7 \\
StoryCloze          & \textbf{95.3} & 86.5	& 79.1 & 94.1 & 53.0 & 82.3 & \textbf{83.5} & 57.4 & 53.0 & 51.4 \\
WinoGrande          & 50.6 & 50.0 & 49.2 & 49.2 & \textbf{50.8} & 50.1 & \textbf{50.8} & 50.1 & \textbf{50.8} & 50.4\\
WSC                 & 39.4 & \textbf{50.0} & 47.1 & 46.2 & 36.3 & 36.5 & \textbf{42.3} & 36.5 & 36.3 & 63.5\\
COPA                & \textbf{95.0} & 74.0	& 80.0 & 88.0 & 60.0 & 79.0 & \textbf{66.0} & 44.0 & 60.0 & 55.0 \\
WiC                 & 54.9 & 51.1 & 51.4 & \textbf{57.5} & 51.7 & 51.9 & 49.4 & 50.0 & \textbf{51.7} & 50.0 \\ \midrule
Average             & \textbf{51.3} & 46.5 & 45.9 & 50.4 & 35.9 & 45.7 & \textbf{41.4} & 37.0 & 35.3 & - \\ \bottomrule
\end{tabular}}
\caption{Performance of XL (3B) and base size ($\sim$250 million) models across datasets. `Rand' refers to performance of models trained on randomly chosen P3 subsets of equivalent size to the ones chosen by DEFT, with `Rand-bal' using uniform random sampling across tasks for subset selection. `T5' refers to performance of a non-finetuned T5 model. `BM25' refers to models trained on subsets of equivalent size to DEFT subsets from P3 retrieved using BM25. DROP and QasperEv. Results are F1 scores, CaseHold micro F1, all else accuracy.} 
\label{tab:t5_xl_deft}
\end{table*}

We first assume we have access \textit{only} to unlabeled task-specific data and cannot train on any target task labeled data. We sample 1000 unlabeled instances per dataset and retrieve the 500 nearest neighbors\footnote{We retrieve 2500 nearest neighbours for T5-base as more retrieved neighbors led to better performance.} of each instance. We then train dataset-specific models on each of the retrieved sets. As seen in Table \ref{tab:t5_xl_deft}, our DEFT-XL models generally outperform\footnote{The exceptions WSC and RTE have small evaluation sets and large variance (see Appendix \ref{app:few_shot_res}), leading us to believe these differences are not significant.} T0-3B and other baselines, with a median relative improvement of 13\% over T0-3B. We also see that base-sized models also improve over baselines in Table \ref{tab:t5_xl_deft}---the DEFT-base models have a median relative improvement of 8\% over the random baseline. All DEFT models are trained on subsets of P3 consisting of 0.1--2\% of all P3 data. This confirms our hypothesis that training on a well-chosen subset of P3 is more beneficial for target task performance than training on a uniform sample of all available data. We also note that using dense representations appears crucial, as using BM25 for retrieval underperforms most baselines. Our results suggest that a general language model encoder can still retrieve relevant cross-task neighbors, contrary to the claims made by \citet{Lin2022UnsupervisedCG}.

Remarkably, DEFT-XL outperforms the majority baselines on two target datasets (QasperEvidence, ANLI R2) where T0-3B does not, and DEFT-base on one (COPA). This observation further confirms that multitask models trained on uniformly sampled data might be suffering from negative interference between tasks.


We run a similar experiment with SuperNaturalInstructions \citep[SNI;][]{niv2}, a recent instruction-tuning dataset, as our pool of multitask data\footnote{We use the split of SNI used by \citet{niv2} with only 100 train samples per task as our underlying pool for fair comparison with Tk-Instruct.} and evaluate on a set of 12 diverse held-out test tasks. We use the same pool of data used to train Tk-Instruct \citep{niv2}, which consists of 100 examples from each English-language task in SNI. Notably, this means that DEFT has a much smaller pool of data to retrieve over compared to P3 (75K vs.~100M examples). We find in Table~\ref{tab:niv2_exp_simple} that DEFT models are able to achieve performance similar to a model trained on all data, with each DEFT model only trained on 5\% of the total available data. DEFT models also significantly outperform training on randomly-chosen subsets. See Appendix~\ref{niv2_experiments} for more details.

\begin{table}[]
\centering
\begin{adjustbox}{max width=.45\textwidth}{
\begin{tabular}{@{}lcc@{}}
\toprule
\textbf{Model} & \textbf{Avg. RougeL} & \textbf{Avg. \# Training Samples}\\ \midrule
DEFT-XL & 49.2 & 3523 \\
Rand-XL & 45.7 & 3523 \\
Tk-Instruct & \textbf{50.7} & 75317 \\\bottomrule
\end{tabular}}
\end{adjustbox}
\caption{Performance of models over 12 held-out tasks from SNI. Models are trained on data retrieved from SNI (DEFT, Rand), or all SNI data (Tk-Instruct).}
\label{tab:niv2_exp_simple}
\end{table}

\subsection{Few-shot Finetuning of DEFT Models}
\label{sec:few_shot_exp}

\begin{table*}[]
\centering
\begin{adjustbox}{max width=\textwidth}{
\begin{tabular}{@{}lccccc:c@{}}
\toprule
           & \textbf{T0-3B+IA3} & \textbf{T5+IA3} & \textbf{Rand+IA3} & \textbf{Rand-Bal+IA3} & \textbf{DEFT-Few (1kQ)} & \textbf{DEFT-Few (20-70Q)}      \\ \midrule
RTE        & $77.5_{{2.0}}$ & $57.0_{{4.3}}$ & $\mathbf{83.3_{{1.1}}}$ & $82.9_{{1.0}}$ & $79.4_{{1.3}}$ & $81.3_{{1.6}}$ \\ 
ANLI R1    & $44.9_{{3.0}}$ & $39.6_{{1.8}}$ & $43.3_{{2.3}}$ & $46.5_{{0.9}}$ & $\mathbf{47.3_{{1.4}}}$ & $\mathbf{47.3_{{1.5}}}$ \\
ANLI R2    & $39.5_{{1.7}}$ & $36.5_{{1.4}}$ & $40.3_{{1.6}}$ & $\mathbf{42.9_{{1.8}}}$ & $40.8_{{2.8}}$ & $42.2_{{2.7}}$ \\
ANLI R3    & $40.2_{{2.2}}$ & $34.8_{{1.1}}$ & $39.3_{{2.3}}$ & $\mathbf{44.3_{{2.1}}}$ & $\mathbf{44.3_{{2.1}}}$ & $42.9_{{1.8}}$ \\
CB         & $78.9_{{3.9}}$ & $67.9_{{2.5}}$ & $81.4_{{3.5}}$ & $81.4_{{2.0}}$ & $82.5_{{2.6}}$ & $\mathbf{84.6_{{4.3}}}$ \\
HellaSwag  & $34.7_{{0.6}}$ & $27.5_{{1.1}}$ & $38.1_{{1.1}}$ & $42.1_{{1.6}}$ & $42.5_{{2.1}}$ & $\mathbf{45.9_{{1.8}}}$ \\
StoryCloze & $93.0_{{0.6}}$ & $83.0_{{3.1}}$ & $92.6_{{0.8}}$ & $95.7_{{0.3}}$ & $96.2_{{0.2}}$ & $\mathbf{96.5_{{0.2}}}$ \\
WinoGrande & $50.6_{{1.3}}$ & $49.8_{{0.8}}$ & $51.4_{{2.3}}$ & $54.0_{{2.6}}$ & $\mathbf{55.9_{{3.0}}}$ & $55.2_{{3.1}}$ \\
WSC        & $\mathbf{64.8_{{3.5}}}$ & $51.0_{{1.0}}$ & $55.8_{{3.0}}$ & $61.5_{{5.3}}$ & $63.3_{{5.2}}$ & $59.6_{{3.8}}$ \\
COPA       & $82.0_{{2.7}}$ & $61.6_{{4.2}}$ & $86.6_{{1.7}}$ & $91.4_{{3.0}}$ & $\mathbf{95.4_{{1.5}}}$ & $92.6_{{2.2}}$ \\
WiC        & $54.9_{{1.9}}$ & $56.6_{{3.0}}$ & $54.5_{{2.4}}$ & $56.2_{{2.2}}$ & $\mathbf{57.7_{{2.9}}}$ & $57.4_{{2.9}}$ \\
\midrule
Average    & 60.1 & 51.4 & 60.6 & 63.6 & \textbf{64.1} & \textbf{64.1}\\ 
\bottomrule
\end{tabular}}
\end{adjustbox}
\caption{Performance of IA3 few-shot finetuned models using XL-size checkpoints. For all models we report the mean over 5 runs with the standard deviation as subscript. We report performance for DEFT-Few models using 1000 unlabeled queries (`1kQ') and few-shot queries (`20-70Q'). We find both iterations of DEFT-Few perform statistically significantly better ($p < 0.5$) than all baselines. See section \ref{sec:few_shot_exp} for details.}
\label{tab:ia3_results}
\end{table*}

Next, we assume we are able to label a small number of task-specific examples, and further train our DEFT models. We reuse the XL-size models trained in Section \ref{sec:exp_cross} and further train them using the parameter-efficient IA3 on the few-shot data used by \citet{liu2022few}. As seen in table \ref{tab:ia3_results}, DEFT models with few-shot finetuning (`DEFT-Few (1kQ)') perform on average 7\% better than T0-3B models with few-shot finetuning (`T0-3B+IA3'), with statistically significant gains on 5 datasets. This shows that DEFT models serve as better starting points for few-shot finetuning than T0-3B, providing similar or better performance across all datasets despite being exposed to much less training data. Notably, DEFT-Few significantly outperforms T0-3B+IA3 on WinoGrande, for which zero-shot DEFT did not significantly outperform zero-shot T0-3B. These results suggest DEFT models are more amenable to few-shot fine-tuning than T0-3B. We also find that DEFT-few performs statistically significantly better than the strong Rand-Bal baseline with few-shot finetuning, further highlighting that DEFT is preferable for both zero and few-shot settings.

\paragraph{Few-shot retrieval} In this experiment, we evaluate DEFT in a setting where we have access only to a small number of target-task labeled examples (exactly what is available to T0-3B+IA3), and no additional unlabeled examples. We construct 5 few-shot sets for each dataset, for each set retrieve cross-task neighbors using the few-shot data, finetune T5 models on the retrieved data, and then finally finetune using IA3 on the labeled few-shot data itself. To make up for the smaller query set, we retrieve the closest 2000 neighbors per query instance. As seen in Table~\ref{tab:ia3_results}, this still results in a model that outperforms T0-3B with few-shot tuning (`DEFT-Few (20-70Q)'), and overall achieves similar performance to DEFT-Few (1kQ). Crucially, this shows that DEFT followed by few-shot finetuning may be a better alternative to few-shot finetuning T0-3B even when both methods have \textit{exactly} the same target-task information available.

\section{Analysis}

\subsection{Cross-Task Retrieval}
\label{sec:heatmap}
\paragraph{What gets retrieved?} We analyse what source datasets get selected during retrieval for each evaluation dataset (see Appendix \ref{sec:retrieved_data}, Figure \ref{fig:retrieval_analysis}). We find that for most target datasets, the majority of source datasets are \textit{not} selected, further strengthening our hypothesis that much of the massive multitask pool is not relevant to a given target task, and no single mixture of datasets is optimal for all target tasks. We additionally find that no more than 27\% of all instances within any source dataset is retrieved, suggesting that our approach is also effective at finding relevant subsets of data \textit{within} large datasets.

\begin{figure*}[h]
    \centering
    \adjustbox{max width=\textwidth}{
        \includegraphics{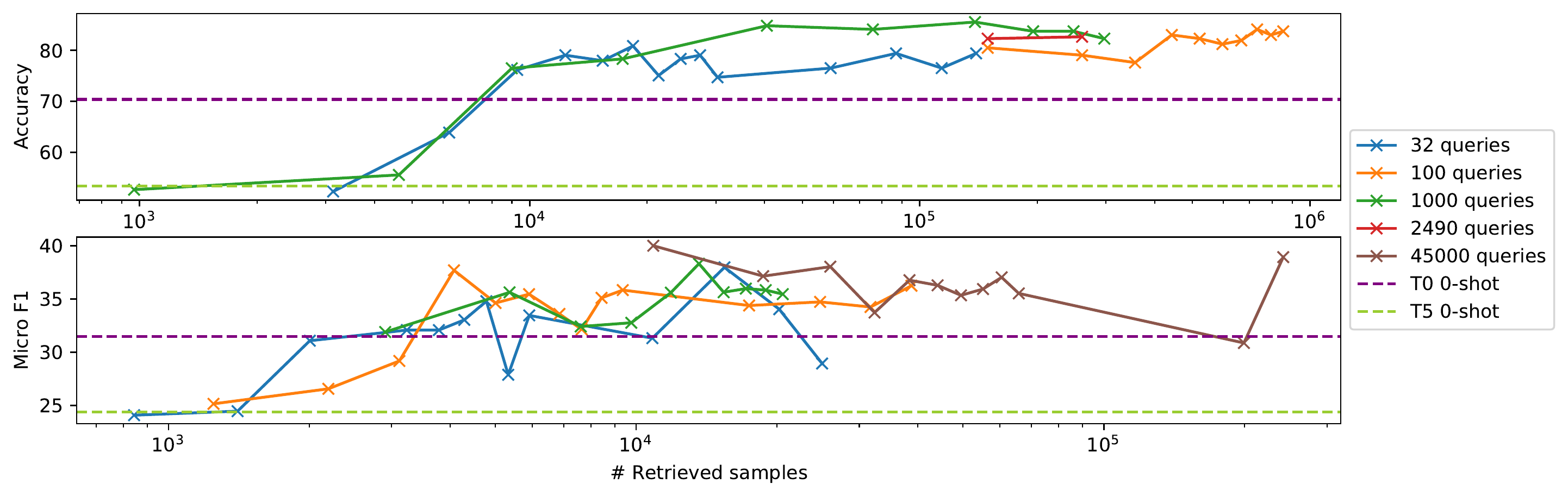}
    }
    \caption{RTE accuracy (above) and CaseHold micro F1 (below) by number of P3 samples retrieved for DEFT-XL across a varying number of neighbors.}
    \label{fig:casehold_sweep}
\end{figure*}

\paragraph{Retrieval hyperparameters} When retrieving cross-task data, the amount and quality of data retrieved is highly dependent on the \textit{query size} (i.e., the number of task-specific instances used for retrieval) and \textit{number of neighbors} (i.e., the number of cross-task samples retrieved per task-specific instance). In Figure \ref{fig:casehold_sweep}, we show the effect of varying both query size (sweeping from 32 to all training data) and the number of neighbors (sweeping from 1 to 5000) on dataset performance on RTE and CaseHold. We find that increasing the amount of data retrieved, whether through increasing the number of neighbors or query set size, results in improved performance up to a point, and then either plateaus or decreases, providing evidence for our hypothesis that using `too much' data can result in reduced downstream performance due to negative interference.

\begin{figure*}[h]
    \centering
    \adjustbox{max width=\textwidth}{
        \includegraphics{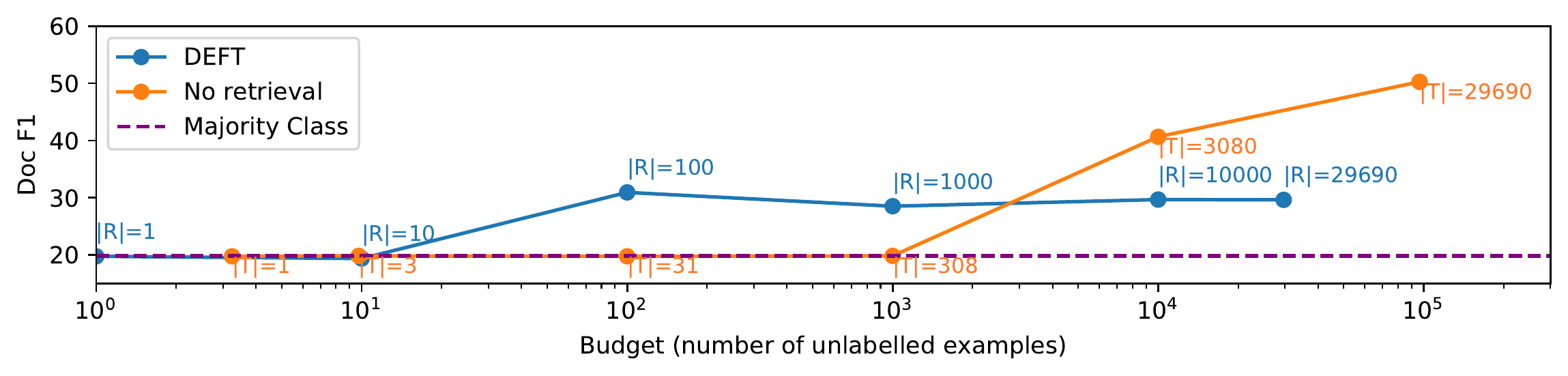}
    }
    \caption{Performance of DEFT-XL and full-finetuning methods with the same annotation budget used for obtaining either labeled or unlabeled data for QasperEvidence. Unless one has a large annotation budget, collecting unlabeled examples is superior to collecting labelled ones. $|R|$ and $|T|$ refer to the size of the retrieval and train sets respectively. See Section \ref{sec:case_study} for details.}
    \label{fig:case_study_results}
\end{figure*}

\paragraph{What model should you use for retrieval?} To determine the effect of of model size on indexing and retrieval, we train models using the cross-task neighbors retrieved by base and XL-size models when the query size and number of neighbors are held constant. We find that using a larger (XL size) indexing model generally results in better performance, but this gap is much larger when training a base size model (8\%) than when training XL-size models (1\%), suggesting that smaller models benefit more from larger retrieval models. We provide detailed results in Appendix \ref{app:index_model_size}.

\paragraph{Are prompts useful for retrieval?} All P3 data is in a prompted format, where the input is made up of (a) the input instance and (b) a prompt that contains information about the task. Training on prompted data greatly aids zero-shot generalisation \citep{wei-etal-2021-shot, sanh2021multitask}, but it is unclear how useful it is for retrieval. To examine this, we run experiments using SuperNaturalInstructions. We index and retrieve the data with and without instructions in the input and compare the performance after training on retrieved subsets.\footnote{We add instructions back into samples without them in order to isolate the effect of instructions on retrieval separate from their effect during finetuning.} We find that retrieving \textbf{without} instructions outperforms retrieving with instructions by a small margin, suggesting that DEFT relies more on instance information rather than task information for retrieval. We provide details in Appendix \ref{niv2_experiments}.

\subsection{Practicality of Assuming Access to Unlabeled Data}
\label{sec:case_study}

Contrary to prior work, our approach assumes access to unlabeled data. This is a practical assumption given that unlabeled data is often readily available or is far cheaper to acquire than labeled data. This is especially true for tasks such as Qasper or CaseHold, which require experts to carefully read (sometimes quite long) texts to provide labels. We argue that DEFT's use of unlabeled data can make it a cost-efficient method to obtain a well-performing task-specific model when the data labeling budget is limited.

We examine this by studying a scenario where QasperEvidence data was collected and assume we have access to P3 and DEFT to make efficient use of it. Obtaining labeled instances for QasperEvidence cost 3.25 times acquiring unlabeled (question-paragraph) instances.\footnote{Based on an estimate provided by the authors of the dataset. Questions were written after reading paper abstracts, and evidence selection required reading entire papers.} We compare (Figure~\ref{fig:case_study_results}) performance on the test set of a T5-XL model trained on a varying number of labeled instances with a DEFT-XL model trained on cross-task nearest neighbors of 3.25 as many unlabeled instances. DEFT yields better results for smaller annotation budgets ($< 1000$ labelled examples), and underperforms models trained on thousands of labelled examples. This confirms our suggestion that DEFT is preferable to regular finetuning for limited data budgets. We also note the DEFT setup makes it easy to use target-task labeled data when available, as shown in Section \ref{sec:few_shot_exp}.



\section{Conclusion}

In this work, we propose Data-Efficient FineTuning, a novel method for efficiently using multitask data by training task-specific models using only a small amount of unlabeled target task data. We use the unlabeled data to select subsets of the multitask data, and train models on these subsets. Our approach performs strongly even when as few as only 20 unlabeled examples are available, and is more effective than full-finetuning on labelled data when it is expensive to gather labelled data, or few (< 3000) labelled data points are available. DEFT models can outperform same-sized models trained on all available data (e.g., T0), despite being trained on significantly less data. Overall, our results strongly suggest that training on all available data, even with large models, is not always the optimal choice and that focusing on ways to better curate higher-quality, smaller datasets is a better path forward.

\section*{Limitations}

Our approach is based on the assumption of a limited data budget, and the observation that general multi-task training may not be the most efficient method when one cares about single target tasks. As such, DEFT is not applicable to ``true'' zero-shot settings where one has no information about the target task, since it relies on the existence of at least some unlabelled examples. Furthermore, for some tasks it may be possible to cheaply gather many examples for finetuning beyond the point where DEFT is useful. In some cases, gathering unlabelled examples may not be so much cheaper than gathering labelled examples that it is worth considering whether to gather unlabelled or labelled examples. Additionally, the recent rise of sparse mixture-of-expert models \citep{shazeer2017, switch} may reduce the negative interference effect observed throughout our work, where DEFT models often outperform models trained on all multitask data and random subsets of the multitask data. Finally, we note that in pilot experiments we found that task diversity was a key element of strong held-out task performance. However, DEFT does not explicitly correct for task diversity, and we leave further exploration for extending DEFT to account for this to future work.

\section*{Ethics Statement}
We believe that the impact of our work is largely positive, showing a case where we are able to achieve good results with significant reductions in the amount of data used to train a model. We hope that this encourages future work in \textit{data-efficiency}, where we attempt to reduce the amount of data required to train an effective NLP model. Such research could aid in making the analysis of the data used to train models easier and cheaper, and reduce the training time and associated carbon cost \citep{energy_nlp} of models. However, we note also that our work currently assumes access to a large pool of multitask data, making it data-efficient only when it comes to training models, and relies on large language models already pretrained over massive datasets.

\bibliography{anthology,iclr2023_conference}

\begin{thebibliography}{56}
\expandafter\ifx\csname natexlab\endcsname\relax\def\natexlab#1{#1}\fi

\bibitem[{Aribandi et~al.(2022)Aribandi, Tay, Schuster, Rao, Zheng, Mehta,
  Zhuang, Tran, Bahri, Ni, Gupta, Hui, Ruder, and Metzler}]{aribandi2022ext}
Vamsi Aribandi, Yi~Tay, Tal Schuster, Jinfeng Rao, Huaixiu~Steven Zheng,
  Sanket~Vaibhav Mehta, Honglei Zhuang, Vinh~Q. Tran, Dara Bahri, Jianmo Ni,
  Jai Gupta, Kai Hui, Sebastian Ruder, and Donald Metzler. 2022.
\newblock \href {https://openreview.net/forum?id=Vzh1BFUCiIX} {Ext5: Towards
  extreme multi-task scaling for transfer learning}.
\newblock In \emph{International Conference on Learning Representations}.

\bibitem[{Bar~Haim et~al.(2006)Bar~Haim, Dagan, Dolan, Ferro, Giampiccolo,
  Magnini, and Szpektor}]{bar2006second}
Roy Bar~Haim, Ido Dagan, Bill Dolan, Lisa Ferro, Danilo Giampiccolo, Bernardo
  Magnini, and Idan Szpektor. 2006.
\newblock The second {PASCAL} recognising textual entailment challenge.

\bibitem[{Bentivogli et~al.(2009)Bentivogli, Dagan, Dang, Giampiccolo, and
  Magnini}]{bentivogli2009fifth}
Luisa Bentivogli, Ido Dagan, Hoa~Trang Dang, Danilo Giampiccolo, and Bernardo
  Magnini. 2009.
\newblock The fifth {PASCAL} recognizing textual entailment challenge.

\bibitem[{Brown et~al.(2020)Brown, Mann, Ryder, Subbiah, Kaplan, Dhariwal,
  Neelakantan, Shyam, Sastry, Askell, Agarwal, Herbert-Voss, Krueger, Henighan,
  Child, Ramesh, Ziegler, Wu, Winter, Hesse, Chen, Sigler, Litwin, Gray, Chess,
  Clark, Berner, McCandlish, Radford, Sutskever, and Amodei}]{brown_gpt_3}
Tom Brown, Benjamin Mann, Nick Ryder, Melanie Subbiah, Jared~D Kaplan, Prafulla
  Dhariwal, Arvind Neelakantan, Pranav Shyam, Girish Sastry, Amanda Askell,
  Sandhini Agarwal, Ariel Herbert-Voss, Gretchen Krueger, Tom Henighan, Rewon
  Child, Aditya Ramesh, Daniel Ziegler, Jeffrey Wu, Clemens Winter, Chris
  Hesse, Mark Chen, Eric Sigler, Mateusz Litwin, Scott Gray, Benjamin Chess,
  Jack Clark, Christopher Berner, Sam McCandlish, Alec Radford, Ilya Sutskever,
  and Dario Amodei. 2020.
\newblock \href
  {https://proceedings.neurips.cc/paper/2020/file/1457c0d6bfcb4967418bfb8ac142f64a-Paper.pdf}
  {Language models are few-shot learners}.
\newblock In \emph{Advances in Neural Information Processing Systems},
  volume~33, pages 1877--1901. Curran Associates, Inc.

\bibitem[{Chalkidis et~al.(2022)Chalkidis, Jana, Hartung, Bommarito,
  Androutsopoulos, Katz, and Aletras}]{chalkidis-etal-2021-lexglue}
Ilias Chalkidis, Abhik Jana, Dirk Hartung, Michael Bommarito, Ion
  Androutsopoulos, Daniel~Martin Katz, and Nikolaos Aletras. 2022.
\newblock Lexglue: A benchmark dataset for legal language understanding in
  english.
\newblock In \emph{Proceedings of the 60th Annual Meeting of the Association
  for Computational Linguistics}, Dubln, Ireland.

\bibitem[{Dagan et~al.(2006)Dagan, Glickman, and Magnini}]{dagan2006pascal}
Ido Dagan, Oren Glickman, and Bernardo Magnini. 2006.
\newblock The {PASCAL} recognising textual entailment challenge.
\newblock In \emph{Machine learning challenges. evaluating predictive
  uncertainty, visual object classification, and recognising tectual
  entailment}, pages 177--190. Springer.

\bibitem[{Dasigi et~al.(2021)Dasigi, Lo, Beltagy, Cohan, Smith, and
  Gardner}]{dasigi-etal-2021-dataset}
Pradeep Dasigi, Kyle Lo, Iz~Beltagy, Arman Cohan, Noah~A. Smith, and Matt
  Gardner. 2021.
\newblock \href {https://doi.org/10.18653/v1/2021.naacl-main.365} {A dataset of
  information-seeking questions and answers anchored in research papers}.
\newblock In \emph{Proceedings of the 2021 Conference of the North American
  Chapter of the Association for Computational Linguistics: Human Language
  Technologies}, pages 4599--4610, Online. Association for Computational
  Linguistics.

\bibitem[{De~Marneffe et~al.(2019)De~Marneffe, Simons, and
  Tonhauser}]{demarneffe:cb}
Marie-Catherine De~Marneffe, Mandy Simons, and Judith Tonhauser. 2019.
\newblock {The CommitmentBank}: Investigating projection in naturally occurring
  discourse.
\newblock To appear in proceedings of Sinn und Bedeutung 23. Data can be found
  at https://github.com/mcdm/CommitmentBank/.

\bibitem[{Dua et~al.(2019)Dua, Wang, Dasigi, Stanovsky, Singh, and
  Gardner}]{dua-etal-2019-drop}
Dheeru Dua, Yizhong Wang, Pradeep Dasigi, Gabriel Stanovsky, Sameer Singh, and
  Matt Gardner. 2019.
\newblock \href {https://doi.org/10.18653/v1/N19-1246} {{DROP}: A reading
  comprehension benchmark requiring discrete reasoning over paragraphs}.
\newblock In \emph{Proceedings of the 2019 Conference of the North {A}merican
  Chapter of the Association for Computational Linguistics: Human Language
  Technologies, Volume 1 (Long and Short Papers)}, pages 2368--2378,
  Minneapolis, Minnesota. Association for Computational Linguistics.

\bibitem[{Fedus et~al.(2022)Fedus, Zoph, and Shazeer}]{switch}
William Fedus, Barret Zoph, and Noam Shazeer. 2022.
\newblock \href {http://jmlr.org/papers/v23/21-0998.html} {Switch transformers:
  Scaling to trillion parameter models with simple and efficient sparsity}.
\newblock \emph{Journal of Machine Learning Research}, 23(120):1--39.

\bibitem[{Ge et~al.(2013)Ge, He, Ke, and Sun}]{ge2013optimized}
Tiezheng Ge, Kaiming He, Qifa Ke, and Jian Sun. 2013.
\newblock Optimized product quantization for approximate nearest neighbor
  search.
\newblock In \emph{Proceedings of the IEEE Conference on Computer Vision and
  Pattern Recognition}, pages 2946--2953.

\bibitem[{Giampiccolo et~al.(2007)Giampiccolo, Magnini, Dagan, and
  Dolan}]{giampiccolo2007third}
Danilo Giampiccolo, Bernardo Magnini, Ido Dagan, and Bill Dolan. 2007.
\newblock The third {PASCAL} recognizing textual entailment challenge.
\newblock In \emph{Proceedings of the ACL-PASCAL workshop on textual entailment
  and paraphrasing}, pages 1--9. Association for Computational Linguistics.

\bibitem[{Gururangan et~al.(2020)Gururangan, Marasovi{\'c}, Swayamdipta, Lo,
  Beltagy, Downey, and Smith}]{gururangan-etal-2020-dont}
Suchin Gururangan, Ana Marasovi{\'c}, Swabha Swayamdipta, Kyle Lo, Iz~Beltagy,
  Doug Downey, and Noah~A. Smith. 2020.
\newblock \href {https://doi.org/10.18653/v1/2020.acl-main.740} {Don{'}t stop
  pretraining: Adapt language models to domains and tasks}.
\newblock In \emph{Proceedings of the 58th Annual Meeting of the Association
  for Computational Linguistics}, pages 8342--8360, Online. Association for
  Computational Linguistics.

\bibitem[{Guu et~al.(2020)Guu, Lee, Tung, Pasupat, and
  Chang}]{guu2020retrieval}
Kelvin Guu, Kenton Lee, Zora Tung, Panupong Pasupat, and Mingwei Chang. 2020.
\newblock Retrieval augmented language model pre-training.
\newblock In \emph{International Conference on Machine Learning}, pages
  3929--3938. PMLR.

\bibitem[{Han and Tsvetkov(2022)}]{han2022orca}
Xiaochuang Han and Yulia Tsvetkov. 2022.
\newblock Orca: Interpreting prompted language models via locating supporting
  data evidence in the ocean of pretraining data.
\newblock \emph{arXiv preprint arXiv:2205.12600}.

\bibitem[{Houlsby et~al.(2019)Houlsby, Giurgiu, Jastrzebski, Morrone,
  De~Laroussilhe, Gesmundo, Attariyan, and Gelly}]{houlsby2019parameter}
Neil Houlsby, Andrei Giurgiu, Stanislaw Jastrzebski, Bruna Morrone, Quentin
  De~Laroussilhe, Andrea Gesmundo, Mona Attariyan, and Sylvain Gelly. 2019.
\newblock Parameter-efficient transfer learning for nlp.
\newblock In \emph{International Conference on Machine Learning}, pages
  2790--2799. PMLR.

\bibitem[{Hu et~al.(2021)Hu, Shen, Wallis, Allen-Zhu, Li, Wang, Wang, and
  Chen}]{hu2021lora}
Edward~J Hu, Yelong Shen, Phillip Wallis, Zeyuan Allen-Zhu, Yuanzhi Li, Shean
  Wang, Lu~Wang, and Weizhu Chen. 2021.
\newblock Lora: Low-rank adaptation of large language models.
\newblock \emph{arXiv preprint arXiv:2106.09685}.

\bibitem[{Jegou et~al.(2010)Jegou, Douze, and Schmid}]{jegou2010product}
Herve Jegou, Matthijs Douze, and Cordelia Schmid. 2010.
\newblock Product quantization for nearest neighbor search.
\newblock \emph{IEEE transactions on pattern analysis and machine
  intelligence}, 33(1):117--128.

\bibitem[{Johnson et~al.(2019)Johnson, Douze, and
  J{\'e}gou}]{johnson2019billion}
Jeff Johnson, Matthijs Douze, and Herv{\'e} J{\'e}gou. 2019.
\newblock Billion-scale similarity search with {GPUs}.
\newblock \emph{IEEE Transactions on Big Data}, 7(3):535--547.

\bibitem[{Khandelwal et~al.(2019)Khandelwal, Levy, Jurafsky, Zettlemoyer, and
  Lewis}]{khandelwal2019generalization}
Urvashi Khandelwal, Omer Levy, Dan Jurafsky, Luke Zettlemoyer, and Mike Lewis.
  2019.
\newblock Generalization through memorization: Nearest neighbor language
  models.
\newblock \emph{arXiv preprint arXiv:1911.00172}.

\bibitem[{Khashabi et~al.(2020)Khashabi, Min, Khot, Sabharwal, Tafjord, Clark,
  and Hajishirzi}]{khashabi-etal-2020-unifiedqa}
Daniel Khashabi, Sewon Min, Tushar Khot, Ashish Sabharwal, Oyvind Tafjord,
  Peter Clark, and Hannaneh Hajishirzi. 2020.
\newblock \href {https://doi.org/10.18653/v1/2020.findings-emnlp.171}
  {{UNIFIEDQA}: Crossing format boundaries with a single {QA} system}.
\newblock In \emph{Findings of the Association for Computational Linguistics:
  EMNLP 2020}, pages 1896--1907, Online. Association for Computational
  Linguistics.

\bibitem[{Kingma and Ba(2015)}]{adam}
Diederik~P. Kingma and Jimmy Ba. 2015.
\newblock \href {http://arxiv.org/abs/1412.6980} {Adam: A method for stochastic
  optimization}.
\newblock In \emph{ICLR (Poster)}.

\bibitem[{Koh and Liang(2017)}]{koh2017understanding}
Pang~Wei Koh and Percy Liang. 2017.
\newblock Understanding black-box predictions via influence functions.
\newblock In \emph{International conference on machine learning}, pages
  1885--1894. PMLR.

\bibitem[{Kung et~al.(2021)Kung, Yin, Chen, Yang, and
  Chen}]{kung-etal-2021-efficient}
Po-Nien Kung, Sheng-Siang Yin, Yi-Cheng Chen, Tse-Hsuan Yang, and Yun-Nung
  Chen. 2021.
\newblock \href {https://doi.org/10.18653/v1/2021.emnlp-main.34} {Efficient
  multi-task auxiliary learning: Selecting auxiliary data by feature
  similarity}.
\newblock In \emph{Proceedings of the 2021 Conference on Empirical Methods in
  Natural Language Processing}, pages 416--428, Online and Punta Cana,
  Dominican Republic. Association for Computational Linguistics.

\bibitem[{Lester et~al.(2021)Lester, Al-Rfou, and
  Constant}]{lester-etal-2021-power}
Brian Lester, Rami Al-Rfou, and Noah Constant. 2021.
\newblock \href {https://doi.org/10.18653/v1/2021.emnlp-main.243} {The power of
  scale for parameter-efficient prompt tuning}.
\newblock In \emph{Proceedings of the 2021 Conference on Empirical Methods in
  Natural Language Processing}, pages 3045--3059, Online and Punta Cana,
  Dominican Republic. Association for Computational Linguistics.

\bibitem[{Levesque et~al.(2011)Levesque, Davis, and
  Morgenstern}]{levesque2011winograd}
Hector~J Levesque, Ernest Davis, and Leora Morgenstern. 2011.
\newblock The {W}inograd schema challenge.
\newblock In \emph{{AAAI} Spring Symposium: Logical Formalizations of
  Commonsense Reasoning}, volume~46, page~47.

\bibitem[{Lewis et~al.(2020)Lewis, Perez, Piktus, Petroni, Karpukhin, Goyal,
  K{\"u}ttler, Lewis, Yih, Rockt{\"a}schel et~al.}]{lewis2020retrieval}
Patrick Lewis, Ethan Perez, Aleksandra Piktus, Fabio Petroni, Vladimir
  Karpukhin, Naman Goyal, Heinrich K{\"u}ttler, Mike Lewis, Wen-tau Yih, Tim
  Rockt{\"a}schel, et~al. 2020.
\newblock Retrieval-augmented generation for knowledge-intensive nlp tasks.
\newblock \emph{Advances in Neural Information Processing Systems},
  33:9459--9474.

\bibitem[{Lin et~al.(2022)Lin, Tan, Miller, Tian, and
  Ren}]{Lin2022UnsupervisedCG}
Bill~Yuchen Lin, Kangmin Tan, Chris Miller, Beiwen Tian, and Xiang Ren. 2022.
\newblock Unsupervised cross-task generalization via retrieval augmentation.
\newblock \emph{ArXiv}, abs/2204.07937.

\bibitem[{Lin et~al.(2021)Lin, Ma, Lin, Yang, Pradeep, and
  Nogueira}]{Lin_etal_SIGIR2021_Pyserini}
Jimmy Lin, Xueguang Ma, Sheng-Chieh Lin, Jheng-Hong Yang, Ronak Pradeep, and
  Rodrigo Nogueira. 2021.
\newblock {Pyserini}: A {Python} toolkit for reproducible information retrieval
  research with sparse and dense representations.
\newblock In \emph{Proceedings of the 44th Annual International ACM SIGIR
  Conference on Research and Development in Information Retrieval (SIGIR
  2021)}, pages 2356--2362.

\bibitem[{Liu et~al.(2022)Liu, Tam, Muqeeth, Mohta, Huang, Bansal, and
  Raffel}]{liu2022few}
Haokun Liu, Derek Tam, Mohammed Muqeeth, Jay Mohta, Tenghao Huang, Mohit
  Bansal, and Colin Raffel. 2022.
\newblock Few-shot parameter-efficient fine-tuning is better and cheaper than
  in-context learning.
\newblock \emph{arXiv preprint arXiv:2205.05638}.

\bibitem[{Malkov and Yashunin(2020)}]{hnsw}
Yu~A. Malkov and D.~A. Yashunin. 2020.
\newblock \href {https://doi.org/10.1109/TPAMI.2018.2889473} {Efficient and
  robust approximate nearest neighbor search using hierarchical navigable small
  world graphs}.
\newblock \emph{IEEE Transactions on Pattern Analysis and Machine
  Intelligence}, 42(4):824--836.

\bibitem[{Mishra et~al.(2021)Mishra, Khashabi, Baral, and Hajishirzi}]{natint}
Swaroop Mishra, Daniel Khashabi, Chitta Baral, and Hannaneh Hajishirzi. 2021.
\newblock \href {https://arxiv.org/abs/2104.08773} {Natural instructions:
  Benchmarking generalization to new tasks from natural language instructions}.
\newblock \emph{CoRR}, abs/2104.08773.

\bibitem[{Mostafazadeh et~al.(2017)Mostafazadeh, Roth, Louis, Chambers, and
  Allen}]{mostafazadeh2017lsdsem}
Nasrin Mostafazadeh, Michael Roth, Annie Louis, Nathanael Chambers, and James
  Allen. 2017.
\newblock Lsdsem 2017 shared task: The story cloze test.
\newblock In \emph{Proceedings of the 2nd Workshop on Linking Models of
  Lexical, Sentential and Discourse-level Semantics}, pages 46--51.

\bibitem[{Nie et~al.(2020)Nie, Williams, Dinan, Bansal, Weston, and
  Kiela}]{nie-etal-2020-adversarial}
Yixin Nie, Adina Williams, Emily Dinan, Mohit Bansal, Jason Weston, and Douwe
  Kiela. 2020.
\newblock \href {https://doi.org/10.18653/v1/2020.acl-main.441} {Adversarial
  {NLI}: A new benchmark for natural language understanding}.
\newblock In \emph{Proceedings of the 58th Annual Meeting of the Association
  for Computational Linguistics}, pages 4885--4901, Online. Association for
  Computational Linguistics.

\bibitem[{Pezeshkpour et~al.(2021)Pezeshkpour, Jain, Wallace, and
  Singh}]{pezeshkpour2021empirical}
Pouya Pezeshkpour, Sarthak Jain, Byron~C Wallace, and Sameer Singh. 2021.
\newblock An empirical comparison of instance attribution methods for nlp.
\newblock \emph{arXiv preprint arXiv:2104.04128}.

\bibitem[{Phang et~al.(2018{\natexlab{a}})Phang, F{\'e}vry, and
  Bowman}]{Phang2018SentenceEO}
Jason Phang, Thibault F{\'e}vry, and Samuel~R. Bowman. 2018{\natexlab{a}}.
\newblock Sentence encoders on stilts: Supplementary training on intermediate
  labeled-data tasks.
\newblock \emph{ArXiv}, abs/1811.01088.

\bibitem[{Phang et~al.(2018{\natexlab{b}})Phang, F\'evry, and
  Bowman}]{phang2018stilts}
Jason Phang, Thibault F\'evry, and Samuel~R. Bowman. 2018{\natexlab{b}}.
\newblock Sentence encoders on stilts: Supplementary training on intermediate
  labeled-data tasks.
\newblock \emph{arXiv preprint arXiv:1811.01088v2}.

\bibitem[{Pilehvar and Camacho-Collados(2019)}]{pilehvar2018wic}
Mohammad~Taher Pilehvar and Jose Camacho-Collados. 2019.
\newblock {WiC}: The word-in-context dataset for evaluating context-sensitive
  meaning representations.
\newblock In \emph{Proceedings of NAACL-HLT}.

\bibitem[{Poth et~al.(2021)Poth, Pfeiffer, R{\"u}ckl{\'e}, and
  Gurevych}]{poth-etal-2021-pre}
Clifton Poth, Jonas Pfeiffer, Andreas R{\"u}ckl{\'e}, and Iryna Gurevych. 2021.
\newblock \href {https://doi.org/10.18653/v1/2021.emnlp-main.827} {{W}hat to
  pre-train on? {E}fficient intermediate task selection}.
\newblock In \emph{Proceedings of the 2021 Conference on Empirical Methods in
  Natural Language Processing}, pages 10585--10605, Online and Punta Cana,
  Dominican Republic. Association for Computational Linguistics.

\bibitem[{Pruthi et~al.(2020)Pruthi, Liu, Kale, and
  Sundararajan}]{pruthi2020estimating}
Garima Pruthi, Frederick Liu, Satyen Kale, and Mukund Sundararajan. 2020.
\newblock Estimating training data influence by tracing gradient descent.
\newblock \emph{Advances in Neural Information Processing Systems},
  33:19920--19930.

\bibitem[{Raffel et~al.(2020)Raffel, Shazeer, Roberts, Lee, Narang, Matena,
  Zhou, Li, Liu et~al.}]{raffel2020exploring}
Colin Raffel, Noam Shazeer, Adam Roberts, Katherine Lee, Sharan Narang, Michael
  Matena, Yanqi Zhou, Wei Li, Peter~J Liu, et~al. 2020.
\newblock Exploring the limits of transfer learning with a unified text-to-text
  transformer.
\newblock \emph{Journal of Machine Learning Research}, 21(140):1--67.

\bibitem[{Robertson and Zaragoza(2009)}]{bm25}
Stephen Robertson and Hugo Zaragoza. 2009.
\newblock \href {https://doi.org/10.1561/1500000019} {The probabilistic
  relevance framework: Bm25 and beyond}.
\newblock 3(4):333–389.

\bibitem[{Roemmele et~al.(2011)Roemmele, Bejan, and
  Gordon}]{roemmele2011choice}
Melissa Roemmele, Cosmin~Adrian Bejan, and Andrew~S. Gordon. 2011.
\newblock Choice of plausible alternatives: An evaluation of commonsense causal
  reasoning.
\newblock In \emph{2011 AAAI Spring Symposium Series}.

\bibitem[{Sakaguchi et~al.(2021)Sakaguchi, Bras, Bhagavatula, and
  Choi}]{winogrande}
Keisuke Sakaguchi, Ronan~Le Bras, Chandra Bhagavatula, and Yejin Choi. 2021.
\newblock \href {https://doi.org/10.1145/3474381} {Winogrande: An adversarial
  winograd schema challenge at scale}.
\newblock \emph{Commun. ACM}, 64(9):99–106.

\bibitem[{Sanh et~al.(2021)Sanh, Webson, Raffel, Bach, Sutawika, Alyafeai,
  Chaffin, Stiegler, Scao, Raja et~al.}]{sanh2021multitask}
Victor Sanh, Albert Webson, Colin Raffel, Stephen~H Bach, Lintang Sutawika,
  Zaid Alyafeai, Antoine Chaffin, Arnaud Stiegler, Teven~Le Scao, Arun Raja,
  et~al. 2021.
\newblock Multitask prompted training enables zero-shot task generalization.
\newblock \emph{arXiv preprint arXiv:2110.08207}.

\bibitem[{Shazeer et~al.(2017)Shazeer, Mirhoseini, Maziarz, Davis, Le, Hinton,
  and Dean}]{shazeer2017}
Noam Shazeer, *Azalia Mirhoseini, *Krzysztof Maziarz, Andy Davis, Quoc Le,
  Geoffrey Hinton, and Jeff Dean. 2017.
\newblock \href {https://openreview.net/forum?id=B1ckMDqlg} {Outrageously large
  neural networks: The sparsely-gated mixture-of-experts layer}.
\newblock In \emph{International Conference on Learning Representations}.

\bibitem[{Shi et~al.(2022)Shi, Michael, Gururangan, and
  Zettlemoyer}]{shi2022nearest}
Weijia Shi, Julian Michael, Suchin Gururangan, and Luke Zettlemoyer. 2022.
\newblock Nearest neighbor zero-shot inference.
\newblock \emph{arXiv preprint arXiv:2205.13792}.

\bibitem[{Strubell et~al.(2020)Strubell, Ganesh, and McCallum}]{energy_nlp}
Emma Strubell, Ananya Ganesh, and Andrew McCallum. 2020.
\newblock \href {https://doi.org/10.1609/aaai.v34i09.7123} {Energy and policy
  considerations for modern deep learning research}.
\newblock \emph{Proceedings of the AAAI Conference on Artificial Intelligence},
  34(09):13693--13696.

\bibitem[{Vu et~al.(2020)Vu, Wang, Munkhdalai, Sordoni, Trischler,
  Mattarella-Micke, Maji, and Iyyer}]{vu-etal-2020-exploring}
Tu~Vu, Tong Wang, Tsendsuren Munkhdalai, Alessandro Sordoni, Adam Trischler,
  Andrew Mattarella-Micke, Subhransu Maji, and Mohit Iyyer. 2020.
\newblock \href {https://doi.org/10.18653/v1/2020.emnlp-main.635} {Exploring
  and predicting transferability across {NLP} tasks}.
\newblock In \emph{Proceedings of the 2020 Conference on Empirical Methods in
  Natural Language Processing (EMNLP)}, pages 7882--7926, Online. Association
  for Computational Linguistics.

\bibitem[{Wang et~al.(2019)Wang, Pruksachatkun, Nangia, Singh, Michael, Hill,
  Levy, and Bowman}]{wang2019superglue}
Alex Wang, Yada Pruksachatkun, Nikita Nangia, Amanpreet Singh, Julian Michael,
  Felix Hill, Omer Levy, and Samuel~R. Bowman. 2019.
\newblock Super{GLUE}: A stickier benchmark for general-purpose language
  understanding systems.
\newblock \emph{arXiv preprint 1905.00537}.

\bibitem[{Wang et~al.(2022)Wang, Mishra, Alipoormolabashi, Kordi, Mirzaei,
  Arunkumar, Ashok, Dhanasekaran, Naik, Stap et~al.}]{niv2}
Yizhong Wang, Swaroop Mishra, Pegah Alipoormolabashi, Yeganeh Kordi, Amirreza
  Mirzaei, Anjana Arunkumar, Arjun Ashok, Arut~Selvan Dhanasekaran, Atharva
  Naik, David Stap, et~al. 2022.
\newblock Super-naturalinstructions:generalization via declarative instructions
  on 1600+ tasks.
\newblock In \emph{EMNLP}.

\bibitem[{Wei et~al.(2021{\natexlab{a}})Wei, Bosma, Zhao, Guu, Yu, Lester, Du,
  Dai, and Le}]{wei2021finetuned}
Jason Wei, Maarten Bosma, Vincent~Y Zhao, Kelvin Guu, Adams~Wei Yu, Brian
  Lester, Nan Du, Andrew~M Dai, and Quoc~V Le. 2021{\natexlab{a}}.
\newblock Finetuned language models are zero-shot learners.
\newblock \emph{arXiv preprint arXiv:2109.01652}.

\bibitem[{Wei et~al.(2021{\natexlab{b}})Wei, Huang, Vosoughi, Cheng, and
  Xu}]{wei-etal-2021-shot}
Jason Wei, Chengyu Huang, Soroush Vosoughi, Yu~Cheng, and Shiqi Xu.
  2021{\natexlab{b}}.
\newblock \href {https://doi.org/10.18653/v1/2021.naacl-main.434} {Few-shot
  text classification with triplet networks, data augmentation, and curriculum
  learning}.
\newblock In \emph{Proceedings of the 2021 Conference of the North American
  Chapter of the Association for Computational Linguistics: Human Language
  Technologies}, pages 5493--5500, Online. Association for Computational
  Linguistics.

\bibitem[{Yeh et~al.(2018)Yeh, Kim, Yen, and Ravikumar}]{yeh2018representer}
Chih-Kuan Yeh, Joon Kim, Ian En-Hsu Yen, and Pradeep~K Ravikumar. 2018.
\newblock Representer point selection for explaining deep neural networks.
\newblock \emph{Advances in neural information processing systems}, 31.

\bibitem[{Zellers et~al.(2019)Zellers, Holtzman, Bisk, Farhadi, and
  Choi}]{zellers2019hellaswag}
Rowan Zellers, Ari Holtzman, Yonatan Bisk, Ali Farhadi, and Yejin Choi. 2019.
\newblock Hellaswag: Can a machine really finish your sentence?
\newblock In \emph{Proceedings of the 57th Annual Meeting of the Association
  for Computational Linguistics}.

\bibitem[{Zheng et~al.(2021)Zheng, Guha, Anderson, Henderson, and
  Ho}]{casehold}
Lucia Zheng, Neel Guha, Brandon~R. Anderson, Peter Henderson, and Daniel~E. Ho.
  2021.
\newblock \href {https://doi.org/10.1145/3462757.3466088} {When does
  pretraining help? assessing self-supervised learning for law and the casehold
  dataset of 53,000+ legal holdings}.
\newblock In \emph{Proceedings of the Eighteenth International Conference on
  Artificial Intelligence and Law}, ICAIL '21, page 159–168, New York, NY,
  USA. Association for Computing Machinery.

\end{thebibliography}
\bibliographystyle{acl_natbib}

\appendix

\section{Compute Resources}

We ran all experiments on a server with 8 80GB A100 GPUs. Most models took 7-10 hours to train on a single 80GB A100 GPU.

\section{Dataset Details}
\label{app:dataset_details}
\begin{table*}[t]
\centering
\begin{tabular}{@{}lcccc@{}}
\toprule
\textbf{Dataset} & \textbf{Retrieval} & \textbf{Eval} & \textbf{\#Shots} & \textbf{Retrieval from} \\ \midrule
CaseHold~\citep{casehold} & 1000               & 2900                & -            & Validation                     \\
DROP~\citep{dua-etal-2019-drop} & 1000               & 8535                & -            & Validation                     \\
QasperEvidence~\citep{dasigi-etal-2021-dataset} & 1000              & 43673               & -            & Validation                     \\
RTE* & 1000 & 277 & 32 & Train \\
ANLI R1~\citep{nie-etal-2020-adversarial} & 1000               & 1000                & 50           & Train                          \\
ANLI R2~\citep{nie-etal-2020-adversarial} & 1000               & 1000                & 50           & Train                          \\
ANLI R3~\citep{nie-etal-2020-adversarial} & 1000               & 1000                & 50           & Train                          \\
CB~\citep{demarneffe:cb}& 250                & 56                  & 32           & Train                          \\
HellaSwag~\citep{zellers2019hellaswag}& 1000               & 10003               & 20           & Train                          \\
StoryCloze~\citep{mostafazadeh2017lsdsem}& 1000               & 1871                & 70           & Train                          \\
WinoGrande~\citep{winogrande} & 1000               & 1767                & 50           & Train                          \\
WSC~\citep{levesque2011winograd}& 554                & 104                 & 32           & Train                          \\
COPA~\citep{roemmele2011choice} & 400                & 100                 & 32           & Train                          \\
WiC~\citep{pilehvar2018wic}& 1000               & 638                 & 32           & Train                          \\ \bottomrule
\end{tabular}
\caption{Size of splits used for experiments across datasets. `\#Shots' indicates the number of shots used in few-shot experiments, and `retrieval from' indicates which split we selected retrieval data from. *Following SuperGLUE \cite{wang2019superglue}, RTE data is from RTE 1/2/3/5 \citep{dagan2006pascal, bar2006second, giampiccolo2007third, bentivogli2009fifth}.}
\label{tab:dataset_splits}
\end{table*}

\paragraph{Sizes and Splits} For each dataset used, we provide the number of retrieval and validation examples used in Table \ref{tab:dataset_splits}. We also indicate if the retrieval data was split from the validation or training split. Note any data used to retrieve is held out of the validation split to avoid information leakage. We additionally provide the number of shots used for each dataset. We follow the number of splits used by \citet{liu2022few} and use the data shared by the authors (available at \href{https://github.com/r-three/t-few/tree/master/data/few_shot}{https://github.com/r-three/t-few/tree/master/data/few\_shot}).

\paragraph{Prompts} We list the prompts used for each dataset. \{x\} indicates a space that is filled in by instance data. 
\begin{itemize}
    \item \textbf{CaseHold}: What is the correct holding statement for the following text? Text: \{context\} (A): \{ending 1\} (B): \{ending 2\} (C): \{ending 3\} (D): \{ending 4\} (E): \{ending 5\}
    \item \textbf{DROP}: Passage: \{passage\} Question: \{question\} Answer:
    \item \textbf{QasperEvidence}: Question: \{question\} Paragraph: \{paragraph\} Is the answer to the question in the paragraph? Answer Yes or No.
    \item \textbf{RTE}: \{premise\} Question: Does this imply that ``\{hypothesis\}''? Yes or no?
    \item \textbf{ANLI}: \{premise\} Question: \{hypothesis\} True, False, or Neither?
    \item \textbf{CB}: \{premise\} Question: \{hypothesis\} True, False, or Neither?
    \item \textbf{HellaSwag}: Complete the description with an appropriate ending: First, \{context a\} Then, \{context b\} ... (a) \{ending 1\} (b) \{ending 2\} (c) \{ending 3\} (d) \{ending 4\}
    \item \textbf{StoryCloze}: \{input sentence 1\} \{input sentence 2\} \{input sentence 3\} \{input sentence 4\} What is a possible continuation for the story given the following options ? - \{answer 1\} - \{answer 2\}
    \item \textbf{WinoGrande}: \{sentence\} What does the \_ in the above sentence refer to? \{option1\} or \{option2\}?
    \item \textbf{WSC}: Passage: \{text\} Question: In the passage above, does the pronoun `\{span 1\}' refer to `\{span 2\}'? Answer:
    \item \textbf{COPA}: \{premise\}  As a consequence...  Help me pick the more plausible option: - \{choice 1\} - \{choice 2\}
    \item \textbf{WiC}: \{sentence 1\} \{sentence 2\} Question: Is the word `\{word\}' used in the same sense in the two sentences above? Yes, No?
\end{itemize}

\section{Few-shot Results without IA3}
\label{app:few_shot_res}
For `DEFT-Few (20-70Q)' in Table \ref{tab:ia3_results}, we trained 5 models using DEFT (as we used 5 few-shot sets per dataset). In Table \ref{tab:few_shot_deft} we report the performance of these models \textit{without IA3 training}. Note we did not train few-shot models for CaseHold, QasperEvidence, or DROP, and so do not report results on these datasets. Notably, RTE, CB, and WSC all have quite large standard deviation ($> 3.0$), which suggests our improvements (or deterioration, for WSC) over T0-3B for these datasets may not be significant.

\section{Index Model Size Experiments}
\label{app:index_model_size}
We explored mismatching the index model sizes, training XL size models on cross-task neighbor splits indexed and retrieved using T5-base, and vice-versa. We use a query size of 1000 and retrieve 500 neighbors per query instance. We present the results in Table \ref{tab:index_model_size}.

\section{SuperNaturalInstructions Experiments}
\label{niv2_experiments}

We use version 2.7 of the SuperNaturalInstructions dataset and use the official splits provided, with 100 samples per train and evaluation tasks. This results in a pool of 75,317 train examples. For evaluation, we randomly select one task per evaluation category in Table 5 of \citet{niv2}. Task names are given in Table \ref{tab:niv2_task_list}. We then generate \textbf{two} indices for retrieval: one where each sample is encoded including the task instruction, and one where each sample is encoded without any instruction. We then retrieve using the 100 unlabeled test instances from each chosen evaluation task, matching the format used for the index (i.e., if we retrieve from the index with instructions, we encode our query data with instructions included). In order to isolate the effect of instructions on retrieval, after retrieving examples, we always train on the corresponding examples with instructions included (i.e., when we retrieve examples without using instructions, we add the instructions back into the inputs before finetuning). On average, we retrieve 3.5k training examples, roughly 5\% of the total training data. Additionally, we finetune a T5-XL model using all available training data (`Tk-instruct'), and a random baseline using random subsets of the training data of the same size as the retrieved subsets (`Rand-XL').

We present our results in Table \ref{tab:niv2_exp}. We find that the instruction-augmented and no-instruction retrieval DEFT models achieve similar performance on average, although the no-instruction variant performs slightly higher. Both DEFT models significantly outperform the Rand-XL baseline, suggesting that the retrieval is still effective even when using a large pool of multitask data without instructions or prompts. However, we find that neither DEFT model significantly outperforms Tk-instruct, which we hypothesise is related to the significantly smaller size of SuperNaturalInstructions compared to P3. However, we note that our DEFT-XL models are trained on significantly less data than Tk-instruct, and training all 12 DEFT models is still cheaper than training the Tk-instruct model, using roughly 42,000 examples overall, roughly 56\% of the data used to train Tk-instruct.
\begin{table}[h]
\centering
\adjustbox{max width=\textwidth}{
\begin{tabular}{@{}lc@{}}
\toprule
\textbf{Task}       & \textbf{DEFT-Few (20-70Q)}  \\ \midrule
RTE                 & $73.2_{{4.0}}$ \\
ANLI R1             & $36.1_{{3.0}}$ \\
ANLI R2             & $34.1_{{0.9}}$ \\
ANLI R3             & $40.6_{{2.0}}$ \\
CB                  & $58.2_{{10.5}}$ \\
HellaSwag           & $34.1_{{0.7}}$ \\
StoryCloze          & $95.1_{{0.3}}$ \\
WinoGrande          & $50.6_{{1.2}}$ \\
WSC                 & $51.0_{{5.1}}$ \\
COPA                & $87.8_{{1.1}}$ \\
WiC                 & $50.8_{{1.7}}$ \\ \midrule
Average             & 55.6 \\ \bottomrule
\end{tabular}}
\caption{Performance of XL size models trained using DEFT with few-shot queries. We report the mean and standard deviation over 5 runs.}
\label{tab:few_shot_deft}
\end{table}
\begin{table}[]
\centering
\adjustbox{max width=\textwidth}{
\begin{tabular}{@{}lcc:cc@{}}
\toprule
\textbf{Train Model Size} & \multicolumn{2}{c}{\textbf{Base}} & \multicolumn{2}{c}{\textbf{XL}} \\ \cmidrule(lr){2-3}\cmidrule(lr){4-5}
\textbf{Index Model Size} & \textbf{Base} & \textbf{XL} & \textbf{Base} & \textbf{XL}  \\ \midrule
CaseHold            & 14.8 & \textbf{15.8} & 32.6 & \textbf{37.2} \\
DROP                & 20.8 & \textbf{21.3} & 30.4 & \textbf{31.0} \\
Qasper              & 15.7 & \textbf{18.0} & 23.3 & \textbf{28.5} \\
RTE                 & 53.4 & \textbf{61.7} & \textbf{77.3} & 74.0 \\
ANLI R1             & \textbf{33.3} & \textbf{33.3} & 39.5 & \textbf{39.8} \\
ANLI R2             & \textbf{33.4} & 32.8 & 35.3 & \textbf{37.5} \\
ANLI R3             & 33.2 & \textbf{33.3} & \textbf{42.5} & 41.4 \\
CB                  & \textbf{50.0} & \textbf{50.0} & \textbf{75.0} & 60.7 \\
HellaSwag           & 26.0 & \textbf{27.9} & 31.7 & \textbf{33.1} \\
StoryCloze          & 74.0 & \textbf{76.8} & 94.4 & \textbf{95.3} \\
WinoGrande          & 49.5 & \textbf{50.4} & \textbf{51.4} & 50.6 \\
WSC                 & 41.4 & \textbf{42.3} & \textbf{43.3} & 39.4 \\
COPA                & \textbf{63.0} & 60.0 & 85.0 & \textbf{95.0} \\
WiC                 & \textbf{48.8} & 48.3 & 49.5 & \textbf{54.9} \\ \midrule
Average             & 39.8 & \textbf{42.8} & 50.8 & \textbf{51.3} \\ \bottomrule
\end{tabular}}
\caption{Performance of DEFT models trained on cross-task neighbors retrieved using different-size models.}
\label{tab:index_model_size}
\end{table}

\begin{table*}[]
\centering
\begin{adjustbox}{max width=\textwidth}{
\begin{tabular}{@{}ll@{}}
\toprule
\textbf{Evaluation Category}              & \textbf{Task}      \\ \midrule
Answerability & \texttt{task020\_mctaco\_answerability\_classification} \\
Cause Effect Classification & \texttt{task391\_cod3s\_cause\_effect\_classification} \\
Coreference & \texttt{task1391\_winogrande\_coreference\_resolution}      \\
Data to Text & \texttt{task957\_e2e\_data\_to\_text} \\
Dialogue Act Recognition & \texttt{task879\_schema\_guided\_dstc8\_dialogue\_act\_recognition} \\
Entailment  & \texttt{task937\_defeasible\_nli\_atomic\_textual\_entailment} \\
Grammar Error Correction & \texttt{task1557\_jfleg\_grammar\_error\_correction} \\
Keyword Tagging & \texttt{task613\_liar\_keyword\_tagging} \\
Overlap & \texttt{task039\_qasc\_overlap\_extraction} \\
Question Rewriting & \texttt{task670\_ambigqa\_question\_rewriting} \\
Title Generation & \texttt{task1356\_xlsum\_title\_generation} \\
Word Analogy & \texttt{task1155\_bard\_word\_analogy} \\ \bottomrule
\end{tabular}}
\end{adjustbox}
\caption{List of tasks used for each evaluation category given in Table \ref{tab:niv2_exp}.}
\label{tab:niv2_task_list}
\end{table*}

\begin{table*}[]
\centering
\begin{adjustbox}{max width=\textwidth}{
\begin{tabular}{@{}lcccc@{}}
\toprule
 & \multicolumn{2}{c}{\textbf{DEFT-XL}} & & \\ 
 \cmidrule(r){2-3}\textbf{Evaluation Category} & \textbf{Instr.} & \textbf{No Instr.} & \textbf{Rand} & \textbf{Tk-Instruct} \\ \midrule
Answerability & 48.0 & 48.0 & \textbf{49.0} & 47.0 \\
Cause Effect Classification & 83.3 & 83.3 & 84.7 & \textbf{87.7} \\
Coreference & 61.0 & 51.0 & 43.0 & \textbf{83.0} \\
Data to Text & 34.0 & 34.4 & 33.4 & \textbf{37.9} \\
Dialogue Act Rec. & 65.0 & 61.0 & 59.0 & \textbf{68.0} \\
Entailment & 50.0 & \textbf{68.0} & 13.0 & 19.0 \\
Grammar Error Correction & \textbf{86.3} & 84.8 & 84.7 & 84.8 \\
Keyword Tagging & 17.4 & 17.6 & \textbf{19.2} & 13.3 \\
Overlap & 17.7  & 20.2 & \textbf{22.3} & 17.8 \\
Question Rewriting & 45.8 & 64.0 & 59.9 & \textbf{68.8} \\
Title Generation & \textbf{21.4} & 20.9 & 20.3 & 20.4 \\
Word Analogy & 60.0 & 41.0 & 60.0 & \textbf{61.3} \\\midrule
\textbf{Average} & 49.2 & 49.5 & 45.7 & \textbf{50.7} \\ \bottomrule
\end{tabular}}
\end{adjustbox}
\caption{Performance of XL-size models on 12 tasks from evaluation categories in \citet{niv2}. All results are in RougeL. `Instr.' and `No Instr.' variants of DEFT-XL refer to models trained using subsets of SuperNaturalInstructions that were retrieved using instructions and without using instruction respectively.}
\label{tab:niv2_exp}
\end{table*}

\section{Retrieved Data}
\label{sec:retrieved_data}
\begin{figure*}[]
    \centering
    \adjustbox{max width=\textwidth}{
        \includegraphics{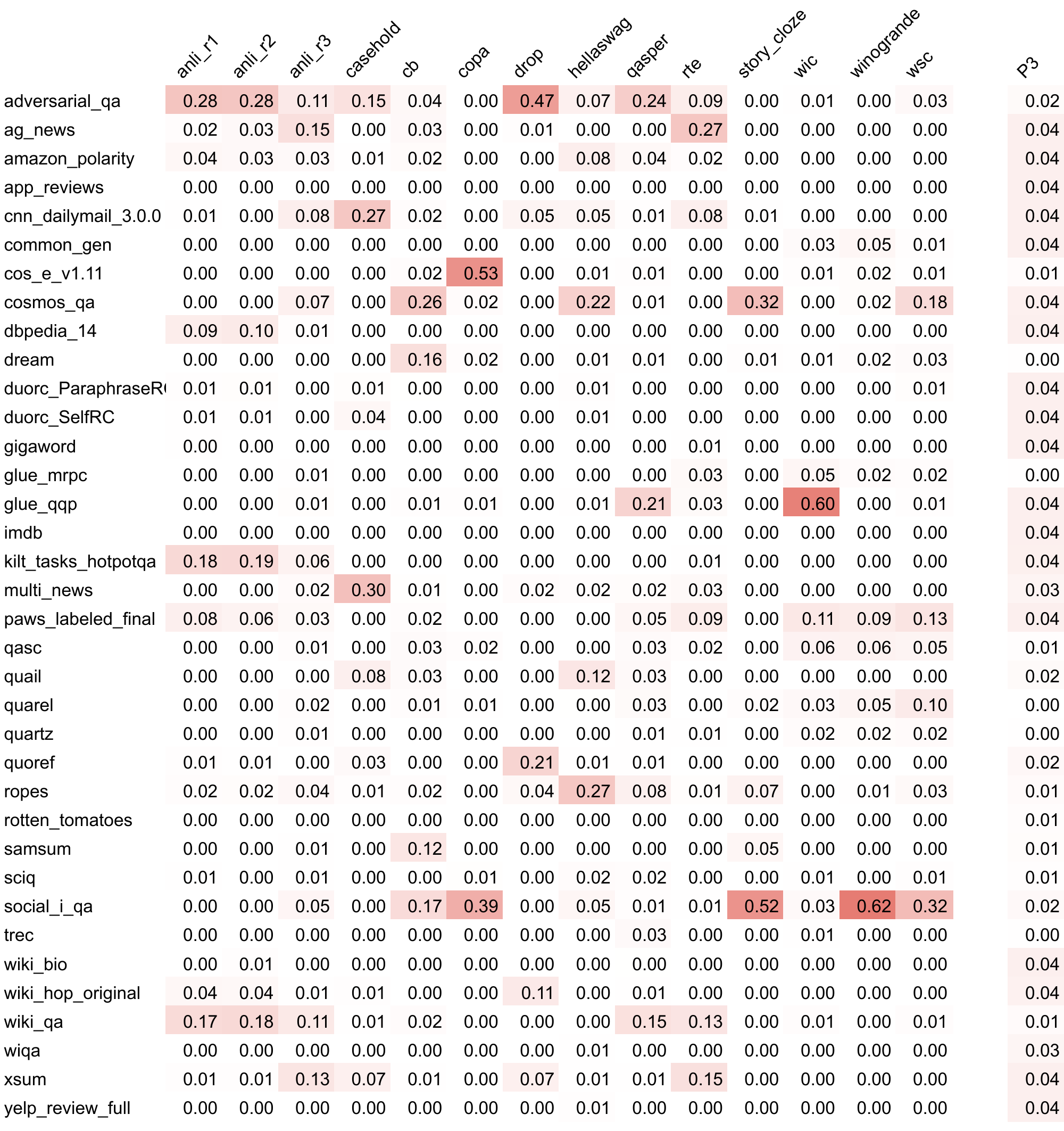}
    }
    \caption{Proportion of the retrieved training data for each evaluation dataset (columns) that comes from each dataset in P3 (rows). The final column shows these values for all of P3.}
    \label{fig:retrieval_analysis}
\end{figure*}

We present a breakdown of the data retrieved for each task using DEFT in Figure~\ref{fig:retrieval_analysis}.

\clearpage
\section{Retrieved Examples}

For a single query from each dataset, we present the top two closest datapoints retrieved below. \textbf{Content warning: some of these datapoints reference sensitive topics.} Queries are chosen randomly. Answers are in \textit{italics}.

\newcommand{\n}{\textbackslash n }
\fbox{\begin{minipage}{.4\textwidth}
\textbf{RTE} 
\vspace{.2em} \hrule \vspace{.2em}
\textbf{Query:} Thanks to a global ban on the ivory trade that was passed in 1989 by the Convention on International Trade in Endangered Species of Wild Fauna and Flora (CITES), the African elephant population may be reversing its spiral toward extinction\textbackslash n Question: Does this imply that "The ban on ivory trade has been effective in protecting the elephant from extinction."? Yes or no?\\ \\
\textbf{Retrieved \#1:} Title: Dissappointed\textbackslash n Review: The software works OK, but haven't gotten any more than three numbers on a draw six lottery after 8 months of trying. The biggest thing to watch out for is support, or lack of. If you rebuild your computer or buy a new one and have to re-install their software, you have to get another product ID from them. It took me almost two weeks of begging and a phone call (just an answering machine on their end) to get a response from them. I am coming up on a week of trying to get a response from them for a product ID for my new computer. Funny, because they responded the next day when I first baught the program and they had my money in hand!\textbackslash n Does this product review convey a negative or positive sentiment? \textit{Negative}\\
\textbf{Retrieved \#2:} You are considering whether to buy a product. You look at the reviews. Would the following review decrease or increase the chances of you buying the product?\textbackslash n Review title: Amazon Rip Off\textbackslash n Product review: What a huge waste of money. I paid \$\$\$ on this very site not but a month ago, now it is \$\$. Got it home, followed the instructions and the silly thing will not get but about a foot off the ground if that, and then it just falls over and beats itself into the ground. Don't waste your cash on this, give your kid a fifty dollar bill and let them light it on fire, they'll have for fun. \textit{decrease}
\end{minipage}}

\fbox{\begin{minipage}{.4\textwidth}
\textbf{ANLI R1}
\vspace{.2em}\hrule\vspace{.2em}
\textbf{Query:} Secrets of the Cryptkeeper’s Haunted House was a children\'s Saturday-morning game show that ran on CBS. It premiered on September 14, 1996 and lasted until August 23, 1997. It featured the Cryptkeeper of "Tales from the Crypt" (with John Kassir as the voice) now serving as an announcer. It is the last TV series in the "Tales From the Crypt" franchise.\textbackslash n Question: The Secrets of the Crypt Keeper\'s House television show aired on CBS until 1997, and then was picked up and aired on NBC for an additional season. True, False, or Neither?\\ \\
\textbf{Retrieved \#1:} Is there a negative or positive tone to this product review?\textbackslash n ===\textbackslash n Title: Not quite as good as some others\textbackslash n Review: This is a fair book, but it is not near as good as Peter O. Steiner's "Thursday Night Poker." Andy Nelson's book can't decide whether it is for beginners or advanced, so it tries to fit advanced technique into too short of space. It barely scratches the surface of any of the topics it brings up. When it doesn't do that, it simply says, "Play so tight that you don't even have to think. Fold 99\% of your hands." That does not make for a fun night, in my opinion.\textbackslash n Answer: \textit{Negative}\\
\textbf{Retrieved \#2:} a delegation from the islamic resistance movement -lrb- hamas -rrb- left the gaza strip monday morning , heading for egypt to hear israel 's response regarding a cairo - mediated ceasefire . In a nutshell, \textit{hamas leaders leave to cairo for final ceasefire discussions}
\end{minipage}}

\fbox{\begin{minipage}{.4\textwidth}
\textbf{ANLI R2}
\vspace{.2em}\hrule\vspace{.2em}
\textbf{Query:} The Sea Wall (French: Un barrage contre le Pacifique ) is a 2008 film by Cambodian director Rithy Panh in a French/Cambodian/Belgian co-production. The film opened on 7 January 2009 in France. It was adapted from the 1950 novel "The Sea Wall" by Marguerite Duras. The novel had previously been adapted as "This Angry Age" by René Clément in 1958.\textbackslash n Question: Marguerite Duras directed the film.  True, False, or Neither?\\ \\
\textbf{Retrieved \#1:} Title: Exactly what I had been looking for!\textbackslash n Review: I've gone through two other iPod FM transmitters that I ended up giving away because the quality was less than desirable. After seeing this one pop up in my Quick Picks last week I decided to give it a try. I used it the very first evening I received it and I'm happy to say my search is over. As others noted, use a low FM frequency for the best results (87.9 in my area works well). I don't receive any interference and the music on my iPod comes through just like I expected. For the price, this is definitely the best deal out there.\textbackslash n Is this product review negative? \textit{No}\\
\textbf{Retrieved \#2:} Based on this review, would the user recommend this product?\textbackslash n===\textbackslash n Review: My friend tried to commit suicide, and while he was bleeding to death, he was watching mtv, and the video for "Hold On" was playing, and he was like "yeah" and after he was done rocking out he got all inspired and called for an ambulance. And now he's still here, and he takes pills that make him tired, and everyone is careful to be very nice to him and be his best friend, even though we all secretly pity him. Thank you so much.\textbackslash n Answer: \textit{No}
\end{minipage}}

\fbox{\begin{minipage}{.4\textwidth}
\textbf{ANLI R3}
\vspace{.2em}\hrule\vspace{.2em}
\textbf{Query:} Well, I think during the campaign, particularly now during this difficult period, we ought to be speaking with one voice, and I appreciate the way the administration has worked hard to calm the tensions. Like the vice president, I call on Chairman Arafat to have his people pull back to make the peace.\textbackslash n Question: Chairman Arafat needs to pull back his people during this difficult time. True, False, or Neither?\\ \\
\textbf{Retrieved \#1:} Title: clinton pushes for greater diversity on wall street\textbackslash n\textbackslash n===\textbackslash n\textbackslash n Write an article with the given title: \textit{u.s. president bill clinton urged wall street brokers to pursue business in america 's economically distressed cities , saying it 's an untapped market with more buying power than mexico .}\\
\textbf{Retrieved \#2:} You are considering whether to buy a product. You look at the reviews. Would the following review decrease or increase the chances of you buying the product?\textbackslash n Review title: Mistake\textbackslash n Product review: I didn't want to "purchase" Bars and Tones". It was a mistake to click on it. This review doesn't deserve so many words.\textbackslash n \textit{decrease}
\end{minipage}}

\fbox{\begin{minipage}{.4\textwidth}
\textbf{WiC}
\vspace{.2em}\hrule\vspace{.2em}
\textbf{Query:} It may rain in which case the picnic will be canceled.\textbackslash n A window case.\textbackslash n Question: Is the word 'case' used in the same sense in the two sentences above? Yes, No?\\ \\
\textbf{Retrieved \#1:} Title: remains of \#\# exhumed from mass graves in eastern croatia\textbackslash n \textbackslash n===\textbackslash n \textbackslash n Write an article with the given title: \textit{thirty bodies believed to be croats killed by ethnic serbs at the outbreak of the \#\#\#\#-\#\# serbo-croatian war in former yugoslavia have been exhumed from two mass graves in eastern croatia , an official said tuesday .}\\
\textbf{Retrieved \#2:} You are considering whether to buy a product. You look at the reviews. Would the following review decrease or increase the chances of you buying the product?\textbackslash n Review title: For the 50-cent table\textbackslash n Product review: My favorite author has run out of steam! His co-author does not, repete, does not have the Paterson style. After sampling this "tandemly"-wriiten book, it becomes obvious that this is a time-waster. Even the editing is bad. I didn't feel guilty about not finishing it. It's headed for the community library's monthly book sale--fifty cent table.\textbackslash n \textit{decrease}
\end{minipage}}

\fbox{\begin{minipage}{.4\textwidth}
\textbf{COPA}
\vspace{.2em}\hrule\vspace{.2em}
\textbf{Query:} The woman filed a restraining order against the man.  As a consequence... \textbackslash n Help me pick the more plausible option:\textbackslash n- The man called her.\textbackslash n- The man stalked her.\\ \\
\textbf{Retrieved \#1:} First sentence of the article: when christopher darden got a recent early-morning call from his publisher that his book `` in contempt '' had become no. \# on the new york times best-seller list , he mumbled something like `` ok , '' then rolled over and went back to sleep .\textbackslash n\textbackslash n Title: \textit{contempt does n't fill christopher darden}\\
\textbf{Retrieved \#2:} "Extract the answer to the following question from the movie plot. If the question isn't answerable, please output "Can't answer".\textbackslash n Question: Who is the toy's leader and Andy's favorite toy?\textbackslash n Title: Toy Story\textbackslash n Movie plot: A boy called Andy Davis (voice: John Morris) uses his toys to act out a bank robbery. The bank is a cardboard box, the robber is Mr. Potato Head (voice: Don Rickles) assisted by Slinky Dog (voice: Jim Varney), and the bystanders include Bo Peep (voice: Annie Potts) and her sheep. The day is saved by cowboy doll Woody (voice: Tom Hanks) playing the sheriff, with help from Rex the dinosaur (voice: Wallace Shawn). Woody is the only toy who gets to say his own lines because he has a pull-string that makes him say things like "Reach for the sky!" and "You're my favorite deputy!"During the opening credits (soundtrack: Randy Newman's "You've Got a Friend in Me"), Andy takes Woody downstairs to find his mother (voice: Laurie Metcalf) decorating the dining room for his birthday party. He asks if they can leave the decorations up until they move, and his mom agrees. She says the guests will arrive soon and sends him back upstairs to get his baby sister Molly (voice: Hannah Unkrich), whose crib is in his room. Andy tosses Woody onto his bed before he pulls Molly out of her crib and carries her away.Woody and the other toys have seemed limp and inanimate up to this point, but as soon as Andy leaves the room, Woody sits up and expresses surprise that the birthday party is today. <cut for space> ...\textbackslash n \textit{Woody}
\end{minipage}}

\fbox{\begin{minipage}{.4\textwidth}
\textbf{WSC}
\vspace{.2em}\hrule\vspace{.2em}
\textbf{Query:} Passage: Dan took the rear seat while Bill claimed the front because his "Dibs!" was quicker. \textbackslash n Question: In the passage above, does the pronoun "his" refer to Dan?\textbackslash n Answer:\\ \\
\textbf{Retrieved \#1:} Title: I want to READ it on my Kindle\textbackslash n Review: Why can't I get the readable version of night for my kindle? I don't want the auidio version...Help! I downloaded it thinking that I would have the choice to read it or to listen to it but that was not the case at all. I'm extremely disappointed.\textbackslash n Does this product review convey a negative or positive sentiment? \textit{Negative}\\
\textbf{Retrieved \#2:} You are considering whether to buy a product. You look at the reviews. Would the following review decrease or increase the chances of you buying the product?\textbackslash n Review title: Look weird - feel great!\textbackslash n Product review: These look so weird and also feel weird when you first put them on but they are so much fun. I love them for my yoga class, and sometimes wear them at night watching TV because the separation they give your toes is good for your feet overall. Try them... you'll become a fan too!\textbackslash n \textit{increase}
\end{minipage}}

\fbox{\begin{minipage}{.4\textwidth}
\textbf{WinoGrande}
\vspace{.2em}\hrule\vspace{.2em}
\textbf{Query:} The phone of Donald is a lot better than Adam's because \_ paid extra for his phone.\textbackslash n What does the \_ in the above sentence refer to? Donald or Adam?\\ \\
\textbf{Retrieved \#1:} Title: more than you expect\textbackslash n Product review: The thing about these tillers is that they do things you might not think about. For instance, they're great for dealing with long-rooted weeds. You can hack your way down to the root, then pull up the plant and not leave a huge hole in the ground.\textbackslash n Would you say this review depicts the product in a flattering or unflattering light?\textbackslash n \textit{flattering}\\
\textbf{Retrieved \#2:} Title: purported statement from al-qaida-linked group says ultimatum against italy ends threatens attacks\textbackslash n\textbackslash n===\textbackslash n\textbackslash n Write an article with the given title: \textit{a statement released sunday in the name of an al-qaida-linked group said the italian government has `` dug its grave by its own hands '' after it ignored a warning to withdraw its troops from iraq by aug. \#\# .}
\end{minipage}}

\fbox{\begin{minipage}{.4\textwidth}
\textbf{HellaSwag}
\vspace{.2em}\hrule\vspace{.2em}
\textbf{Query:} Complete the description with an appropriate ending:\textbackslash n First, [header] How to make a butterfly out of plastic spoons [title] Gather the materials you will need for this project, listed below. [title] Put a craft cloth or some newspaper down on your working surface. [title] Cut the top portion of the four spoons off (leaving about half an inch of the handle left. Then,  ...\\
\textbf{Retrieved \#1:} Title: hmm...\textbackslash n Review: I bought this costume in hopes of wearing for Halloween ( last year). I had even separately purchased the duster ( which I am now using to really dust things). Uhh... I tried it on ( I got a X-Small) and its just big... the net piece ( part of the dress with the dots) go all the way down to almost my knees. Which makes it awkward and not sexy at all- its just weird I tried tucking the net part in to my undies to hold it, but it just becomes supper puffy- again looks weird. I never wore it and its still brand new sitting in my closet somewhere.Maybe its just for my body- I am not sure, but the material isn't as great either compared to the picture. Def. does not look anything close to how the model looks in it.Sorry- this was not a good buy at all. The model sure looks good in it.\textbackslash n Does this product review convey a negative or positive sentiment? \textit{Negative}\\
\textbf{Retrieved \#2:} What type of details about adolf heeb\textbackslash n can be gathered from the following bio?\textbackslash n\textbackslash n Bio: adolf heeb -lrb- born 11 july 1940 -rrb- is a former cyclist and politician from liechtenstein .\textbackslash n he competed in the individual road race at the 1960 summer olympics .\textbackslash n he later served as a member of the landtag of liechtenstein and leader of the patriotic union party.
\end{minipage}}

\fbox{\begin{minipage}{.4\textwidth}
\textbf{CB}
\vspace{.2em}\hrule\vspace{.2em}
\textbf{Query:} B: boy, he's a big one. A: he's pretty big. That's why it really surprises me, you know, that he hasn't come back, because, like I said, he's never gone away like this before, and, I would think, you know, I mean, he might could get hurt by a car or something. I don't know that he could really get killed that easily because he is so big.\textbackslash n Question: he could really get killed that easily True, False, or Neither?\\ \\
\textbf{Retrieved \#1:} Summarize this document: Glen Water Limited also paid costs of \textbackslash u00a31,600 to restore fish stocks in the Tall River near Richhill.\textbackslash n About 250 metres of the river was affected when untreated sewage was discharged into it.\textbackslash n It caused what was described as a \"moderate\" fish kill.\textbackslash n Inspectors found a plume of untreated sewage coming from a discharge pipe at Richhill waste water treatment works in 2014.\textbackslash n An investigation found that an \"uninterruptable power source\" at the plant had failed.\textbackslash n In addition, a power cut to the alarm system meant staff were unaware of the problem.\textbackslash n Glen Water Limited is based at Dartford in Kent.\n Under a 25-year public private partnership it has the contract for 25\% of Northern Ireland's waste water treatment capacity.\n It operates and maintains nine treatment works or pumping stations up to 2032 in return for monthly payments.\n Summary: \textit{A company which treats sewage for NI Water under a public private partnership contract has been fined \textbackslash u00a32,500 for polluting a County Armagh river.}\\
\textbf{Retrieved \#2:} Title: Good\n Review: Well, I'd say all of these songs are well constructed, dope lyrics whatever... but wth? all the basslines sound the same or what? Personally i prefer Violent By Design over this.\n Is this product review negative? \textit{No}
\end{minipage}}

\fbox{\begin{minipage}{.4\textwidth}
\textbf{StoryCloze}
\vspace{.2em}\hrule\vspace{.2em}
\textbf{Query:} Andy had always wanted a big kids bike. When he turned six Year's old he asked for a bike for his birthday. He did not know how to ride a bike. On Andy's birthday his mother gave him a bike. What is a possible continuation for the story given the following options ?\n - Andy cried for hours.\n - His dad taught him how to ride it.\\ \\
\textbf{Retrieved \#1:} Based on this review, would the user recommend this product?\n ===\n Review: I love most Neil Young but every fan knows that about one in three of his albums really sucks. After Greendale and Greatest hits, I'm very disapointed.\n Answer: \textit{No}\\
\textbf{Retrieved \#2:} hong kong share prices rose a mere \#.\#\# percent on late overseas buying thursday despite early profit-taking , dealers said .\n \n ===\n \n Given the above sentence, write its title: \textit{hong kong shares close \#.\#\# percent firmer}
\end{minipage}}

\fbox{\begin{minipage}{.4\textwidth}
\textbf{CaseHOLD}
\vspace{.2em}\hrule\vspace{.2em}
\textbf{Query:} What is the correct holding statement for the following text?\n Text: component of the res judicata doctrine. The Ohio Supreme Court held that the original criminal proceedings in Krahn were insufficient to invoke collateral estoppel in the later malpractice case because the claimed error by Krahn’s criminal lawyer in plea negotiations was not “ ‘actually and necessarily litigated and determined’ in the denial of her motion to vacate the criminal judgment against her.” Krahn, 43 Ohio St.3d at 108, 538 N.E.2d 1058, quoting Goodson v. McDonough Power Equip., Inc. (1983), 2 Ohio St.3d 193, 195, 2 OBR 732, 443 N.E.2d 978. The Supreme Court by no means suggested that collateral estoppel was completely inapplicable in the context of a criminal conviction when, as here, matters genuinely were litigated and determined. Id. at 107, 538 N.E.2d 1058 (<HOLDING>). Decisions in Ohio other than Krahn relative \n (A): recognizing the doctrine of collateral estoppel in agency proceedings\n (B): holding that the facts prevent the invocation of collateral estoppel as a bar to krahns cause of action in this case\n (C): holding collateral estoppel elements met considering changed circumstances in the context of an exception to the general rule of collateral estoppel\n (D): recognizing the cause of action\n (E): holding that collateral estoppel applies to  1983 claims \\ \\
\textbf{Retrieved \#1:} Is there a negative or positive tone to this product review?\n ===\n Title: Too steep\n Review: I bought this for my dog who had back problems, it was way too steep and my dog had to jump about 3/4's of the way up to my bed because the measurement of the ramp on the description was incorrect. It totally defeated the purpose of my dog having to not jump. I had to go back to the stairs I had been using\n Answer: \textit{Negative}\\
\textbf{Retrieved \#2:} Write a title for this sentence: the fate of president barack obama 's top domestic priority -- a remake of the u.s. health care system -- now rests in the hands of a pivotal but deeply divided senate committee . \n\n Title: \textit{toughest test coming up for health care overhaul}
\end{minipage}}

\fbox{\begin{minipage}{.4\textwidth}
\textbf{DROP}
\vspace{.2em}\hrule\vspace{.2em}
\textbf{Query:} Passage:  Coming off their overtime win at San Diego, the Broncos traveled to the Mall of America Field at the Hubert H. Humphrey Metrodome for an interconference duel with the Minnesota Vikings. The game's first points came from the Vikings, when defensive end Jared Allen tackled running back Willis McGahee in the end zone for a safety. The Broncos grabbed the lead when linebacker Mario Haggan returned an interception off Vikings' quarterback Christian Ponder 16 yards for a touchdown ... <cut for space> ... On the Broncos' next possession, McGahee rushed 24 yards for a touchdown and Tebow scrambled for a two-point conversion to tie the game at 29. The Vikings subsequently reclaimed the lead on Longwell's 39-yard field goal with 3:06 left in the game. The Broncos answered with kicker Matt Prater's 46-yard field goal with 1:33 left to tie the game at 32. On the Vikings' ensuing possession, Broncos' cornerback Andr\&\#233; Goodman returned an interception off Ponder to the Vikings' 15-yard line. Six plays later, Prater nailed the game-winning 23-yard field goal as time expired to give the Broncos their fifth consecutive win.\n Question: how many yards did longwell make?\n Answer:\\ \\
\textbf{Retrieved \#1:} Make a title for this article: andy roddick hit a record-breaking \#\#\# mph -lrb- \#\#\#.\# kph -rrb- serve friday in a lopsided win over stefan koubek as the united states took a \#-\# davis cup lead over austria . \textit{\n \n roddick ginepri give united states \#-\# lead over austria}\\
\textbf{Retrieved \#2:} Orton does not start against Ohio State Purdue quarterback Kyle Orton did not start Saturday \#39;s game against Ohio State, though he was listed as available to play. Orton has been bothered by a right hip injury for the last month. \n \n Which of the following sections of a newspaper would this article likely appear in? World News, Sports, Business, or Science and Technology? \textit{Sports}
\end{minipage}}

\fbox{\begin{minipage}{.4\textwidth}
\textbf{Qasper}
\vspace{.2em}\hrule\vspace{.2em}
\textbf{Query:} Question: How big is Augmented LibriSpeech dataset? Paragraph: We introduce a multilingual speech-to-text translation corpus, CoVoST, for 11 languages into English, diversified with over 11,000 speakers and over 60 accents. We also provide baseline results, including, to our knowledge, the first end-to-end many-to-one multilingual model for spoken language translation. CoVoST is free to use with a CC0 license, and the additional Tatoeba evaluation samples are also CC-licensed. Is the answer to the question in the paragraph? Answer Yes or No.\\ \\
\textbf{Retrieved \#1:} Title: make your july \# celebration sizzle\n \n ===\n \n Write an article with the given title: \textit{you have less than a week to get your fourth of july cookout menu set and we thought we 'd help .}\\
\textbf{Retrieved \#2:} Title: A good idea...\n Review: that went terribly bad. I cannot comprehend how some of these "artists" were chosen for this. "Atlantic City" and "State Trooper" are embarrasing to say the least, but they sadly showcase what is now Nashville's finest. If Johnny Cash and Dar Williams recordings had not appeared on this CD, one star would have been too many. Thankfully, these mostly pathetic renderings cannot tarnish the greatness of Mr. Springsteen or his amazing album. Go get the original. You won't be sorry.\n Does this product review convey a negative or positive sentiment? \textit{Negative}
\end{minipage}}
\end{document}